\documentclass{article} 
\usepackage{iclr2026_conference,times}

\usepackage{amsmath,amsfonts,bm}

\def\eqref#1{equation~\ref{#1}}

\def\1{\bm{1}}

\DeclareMathAlphabet{\mathsfit}{\encodingdefault}{\sfdefault}{m}{sl}
\SetMathAlphabet{\mathsfit}{bold}{\encodingdefault}{\sfdefault}{bx}{n}

\usepackage{hyperref}
\usepackage{url}
\usepackage{booktabs,multirow,tabularx,makecell,xcolor}
\usepackage{graphicx}
\usepackage{siunitx}
\usepackage{multirow}
\usepackage{colortbl}
\usepackage{hhline}

\usepackage{booktabs}
\usepackage{xcolor}   
\usepackage{pifont}
\newcommand{\cmark}{\ding{51}}  
\usepackage[ruled,vlined]{algorithm2e}
\hypersetup{
    colorlinks=true,
    linkcolor=red,
    citecolor=cyan,
    }

\usepackage{amsmath,amssymb,amsthm}

\newtheorem{proposition}{Proposition}
\newtheorem{remark}{Remark}

\title{\textsc{Self-Calibrated Consistency} can Fight Back for Adversarial Robustness in Vision-Language Models}

\iclrfinalcopy

\author{%
   Jiaxiang Liu$^{1}$ \quad Jiawei Du$^2$ \quad Xiao Liu$^1$\quad
   Prayag Tiwari$^3$\quad Mingkun Xu$^{1, *}$ 
   \quad\\
   $^1$Guangdong Institute of Intelligence Science and Technology, Hengqin, Zhuhai, China \\
   $^2$Agency for Science, Technology and Research (A*STAR), Singapore \\
   $^3$School of Information Technology, Halmstad University, Halmstad, Sweden \\ \\
}

\begin{document}

\maketitle

\begin{abstract}
Pre-trained vision-language models (VLMs) such as CLIP have demonstrated strong zero-shot capabilities across diverse domains, yet remain highly vulnerable to adversarial perturbations that disrupt image-text alignment and compromise reliability. 
Existing defenses typically rely on adversarial fine-tuning with labeled data, limiting their applicability in zero-shot settings. 
In this work, we identify two key weaknesses of current CLIP adversarial attacks—lack of semantic guidance and vulnerability to view variations—collectively termed semantic and viewpoint fragility. To address these challenges, we propose \textsc{Self-Calibrated Consistency (SCC)}, an effective test-time defense. SCC consists of two complementary modules: 
\textit{Semantic consistency}, which leverages soft pseudo-labels from counterattack warm-up and multi-view predictions to regularize cross-modal alignment and separate the target embedding from confusable negatives; and
\textit{Spatial consistency}, aligning perturbed visual predictions via augmented views to stabilize inference under adversarial perturbations.
Together, these modules form a plug-and-play inference strategy. 
Extensive experiments on 22 benchmarks under diverse attack settings show that SCC consistently improves the zero-shot robustness of CLIP while maintaining accuracy, and can be seamlessly integrated with other VLMs for further gains. 
These findings highlight the great potential of establishing an adversarially robust paradigm from CLIP, with implications extending to broader vision-language domains such as BioMedCLIP. 
\end{abstract}

\section{Introduction}
With the rapid proliferation of image-text data and advances in self-supervised learning, 
vision-language models (VLMs) have attracted increasing attention from both academia and industry \citep{radford2021learning, chen2023vlp, liu2025kpl, wang2025fair}. 
Among them, CLIP has demonstrated impressive zero-shot capabilities, effectively aligning images with descriptive text and enabling strong transfer across classification, retrieval, and diverse downstream tasks \citep{zhou2022extract, shin2022reco,liu2024vpl,zhao2022exploiting, zhang2023simple}. However, recent studies reveal that even subtle, imperceptible perturbations can cause CLIP to misclassify, exposing a fundamental vulnerability shared by many neural networks \citep{radford2021learning}. 
As foundation models are increasingly deployed in real-world applications, ensuring their adversarial robustness has become critical \citep{xing2025clip}. This work investigates the robustness of CLIP and its derivatives under such perturbations.

CLIP, unlike conventional models with well-studied adversarial robustness, is a foundation model pre-trained on massive image–text pairs. It encodes broad real-world knowledge yet requires careful handling to preserve generalization, particularly under adversarial attacks \citep{zhou2022learning,zhou2022conditional}.
Since its pretraining demands large-scale data and substantial computational resources, most practitioners rely on open-source variants from a limited pool of models \citep{zhang2025multimodal}, leaving CLIP-based applications especially exposed to adversarial risks. Recent studies further reveal that VLMs are highly susceptible to such perturbations, undermining their reliability in open-world deployment \citep{li2024one,schlarmann2024robust,malik2025robust}.

Research on CLIP’s adversarial robustness is still nascent. 
A main line of work is training-based defenses, including adversarial fine-tuning (AFT) \citep{malik2025robust,schlarmann2024robust} and adversarial prompt tuning (APT) \citep{shu2022test, zanella2024test}. AFT fine-tunes the visual encoder via a min–max game with dynamically generated adversarial images, yielding transferable zero-shot robustness but at high computational cost, reliance on labeled data, and a tendency to overfit the fine-tuning set, which degrades generalization on unseen distributions. APT instead adjusts learnable tokens in the text embedding space to align adversarial images, but similarly overfits to training data—boosting clean accuracy only on seen distributions while harming generalization \citep{yu2024text}.
Another emerging line is test-time defense, which adapts models during inference without retraining. Recent works include R-TPT \citep{sheng2025r}, minimizing pointwise entropy with reliability-weighted ensembles, and Test-Time Counterattack (TTC) \citep{xing2025clip}, leveraging CLIP’s visual encoder to counter adversarial perturbations. While promising, both remain prone to semantic misalignment and unstable recovery under attacks.

\begin{figure*}[t]
\centering
\begin{minipage}[t]{0.48\textwidth}
    \centering
    \includegraphics[height=0.5\textwidth]{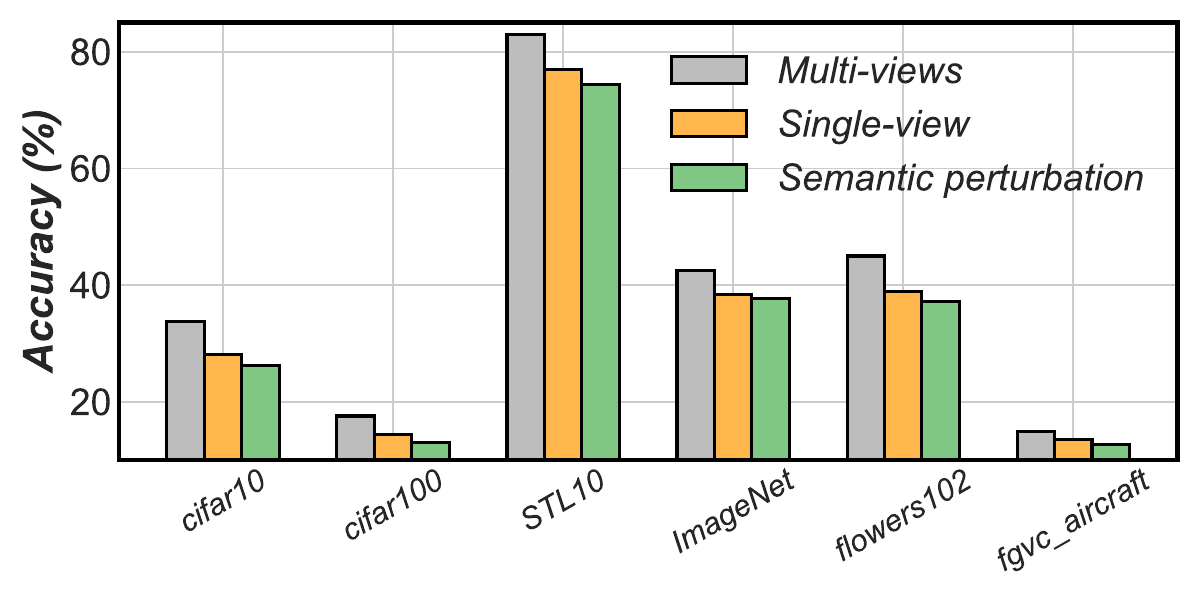}
    \caption{
    Analysis of Counterattack for Adversarial Robustness. Performance drops when reducing from two views to a single view, and degrades further under semantic perturbations.
    }
    \label{analyis-findings}
\end{minipage}%
\hfill
\begin{minipage}[t]{0.48\textwidth}
    \centering
    \includegraphics[height=0.62\textwidth]{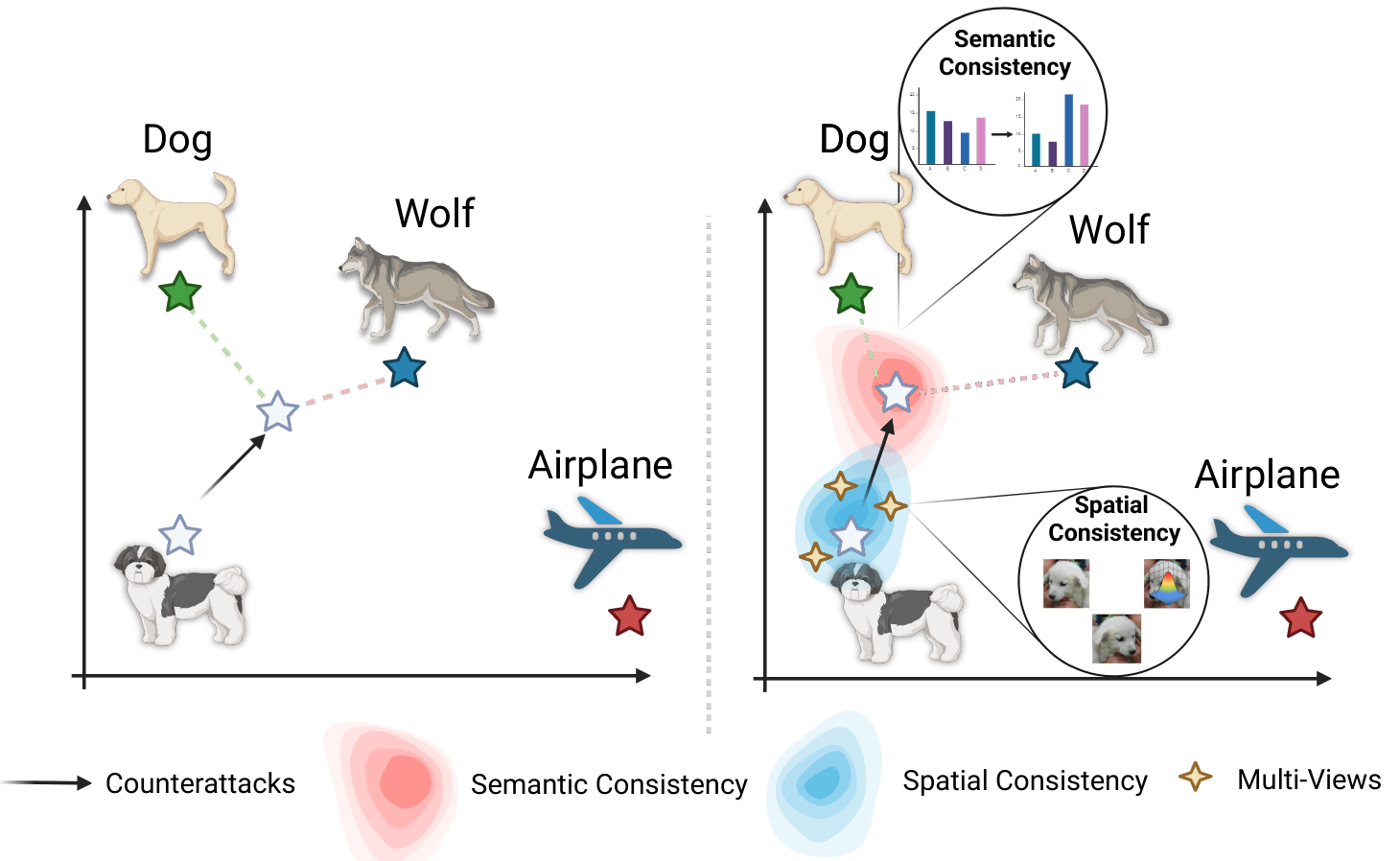}
    \caption{
    During counterattack inference, embeddings tend to drift within the adversarial space and fall into hard-negative traps; SCC leverages cross-modal semantic and spatial consistency to push them away from hard samples and back toward the correct class space.
    }
    \label{dog-wolf}
\end{minipage}
\end{figure*}

Building on prior robustness studies, adversarial attacks often induce pseudo-stability, where perturbed images appear deceptively stable \citep{xing2025clip}; thresholded counter-attacks mitigate this but still shift embeddings toward hard negatives and leave single-view corrections insufficient to suppress noise, as shown in \autoref{analyis-findings}.
Motivated by these observations, we propose Self-Calibrated Consistency (SCC), a simple yet effective test-time defense composed of two complementary components. 
\textit{Semantic consistency}, which leverages
soft pseudo-labels from counterattack warm-up and multi-view predictions to regularize cross-modal alignment and separate target embeddings from confusable negatives (\autoref{dog-wolf}); and
\textit{Spatial consistency}, which enforces agreement among perturbed visual predictions and leverages augmented views to mitigate viewpoint fragility and stabilize feature calibration (\autoref{dog-wolf}).
Extensive experiments on 22 zero-shot benchmarks demonstrate that SCC consistently improves adversarial robustness while preserving clean accuracy, surpassing state-of-the-art test-time defenses.
In summary, our main contributions are:
\begin{itemize}
 \item  
This work uncovers and theoretically analyzes three vulnerabilities in test-time defenses—semantic drift, view sensitivity, and hard-negative dominance—and proposes SCC, a framework that shifts the paradigm from unimodal defenses to cross-modal, multi-view self-corrective robustness.
 
 \item  
SCC unifies semantic and spatial consistency into a principled test-time defense: a cross-modal consistency constraint preserves alignment against hard negatives, while spatial consistency stabilizes perturbed views to mitigate viewpoint fragility, together forming a dual defense that delivers robust and generalizable zero-shot performance.
 \item 
 \textsc{SCC} is a plug-and-play defense that boosts robustness without retraining, consistently outperforming prior test-time methods on 22 benchmarks and extending effectively to CLIP derivatives such as BioMedCLIP.
\end{itemize}

\begin{figure*}[t]
\centering
\includegraphics[width=\textwidth]{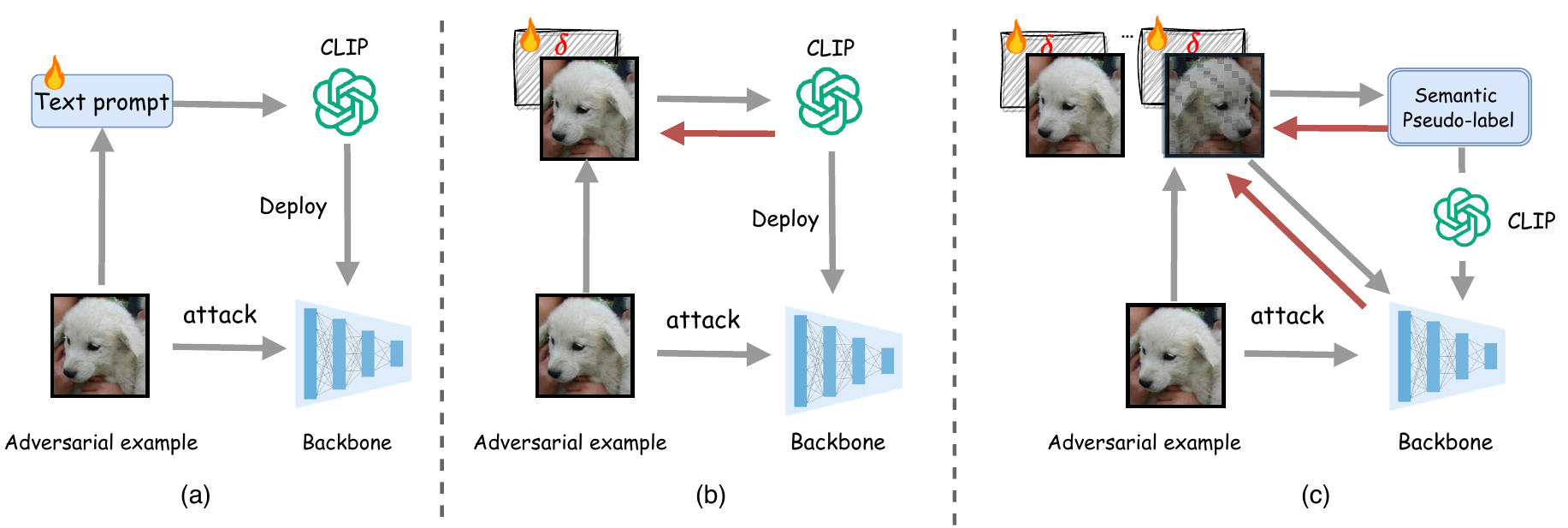}
\caption{
Test-time defense paradigms on CLIP. (a). R-TPT adapts text prompts online but still suffers from adversarial perturbations. (b). TTC repairs adversarial inputs via corrective perturbations, yet remains sensitive to view variance and hard negatives. (c). \textsc{SCC} enforces semantic and spatial consistency, yielding more stable recovery.
} 
\label{Figure_compare}
\end{figure*}

\section{Preliminaries and Related Work}
\label{gen_inst}

Despite notable success, VLMs are highly vulnerable to adversarial perturbations: imperceptible changes crafted by PGD \citep{madry2017towards} or CW can flip predictions, and multimodal misalignment exacerbates this by shifting image embeddings toward hard negatives, causing semantic drift \citep{su2019one,moosavi2017universal,andriushchenko2020square,ilyas2018black}.

To address adversarial vulnerability in VLMs, several directions have been explored \citep{mao2022understanding,li2024one,liang2024comprehensive, yu2023benchmarking, shu2022test}. 
AFT \citep{malik2025robust,schlarmann2024robust} enhances robustness with adversarial examples but is costly, label-dependent, and overfits, hurting zero-shot generalization. 
APT \citep{yu2024text} adjusts learnable tokens in the text space, yet also overfits, inflating clean accuracy only on seen data while degrading unseen performance. 
Test-time defenses, including R-TPT \citep{sheng2025r} and TTC \citep{xing2025clip}, adapt models without retraining but remain unstable and semantically misaligned under attacks \citep{shu2022test,zanella2024test,sui2024just}. 
Overall, existing methods either demand expensive retraining or fail to ensure semantic and stable predictions \citep{yu2024text,abdul2024align}, motivating our SCC (\autoref{Figure_compare}).

\textbf{Problem formulation:}
Given an image $x$ and a set of text prompts $\{t_k\}$, zero-shot classification in CLIP is performed by computing cosine similarities between the normalized image embedding $f_{\text{img}}(x)$ and text embeddings $g_{\text{text}}(t_k)$, followed by a softmax over classes:
$p(y=k \mid x) = \frac{\exp\!\left(\tau \cdot \langle f_{\text{img}}(x), g_{\text{text}}(t_k)\rangle \right)}{\sum_j \exp\!\left(\tau \cdot \langle f_{\text{img}}(x), g_{\text{text}}(t_j)\rangle \right)},$
where $\tau$ denotes a learnable temperature parameter.

We consider CLIP, consisting of an image encoder $f_{\text{img}}(\cdot)$ and a text encoder $g_{\text{text}}(\cdot)$. 
Given an image $x$ and class prompts $\{t_k\}_{k=1}^K$, zero-shot prediction is
\begin{equation}
    \hat{y} = \arg\max_k \;\langle f_{\text{img}}(x), g_{\text{text}}(t_k)\rangle,
\end{equation}

In adversarial settings, an attacker perturbs $x$ within an $\ell_p$ ball of radius $\epsilon_a$ (perturbation budget), yielding
$
    x^{\text{adv}} = x + \delta^{\text{atk}}, \quad \|\delta^{\text{atk}}\|_{p} \le \epsilon_a,
$
To counteract this, our defense applies a corrective perturbation $\delta$ to recover alignment:
\begin{equation}
    x^{\text{cnt}} = x^{\text{adv}} + \delta, \quad \|\delta\|_{p} \le \epsilon_d,
\end{equation}
Here, $\delta$ is optimized at test time, and $\epsilon_d$ controls the maximum allowable perturbation magnitude.

\begin{figure*}[t]
\centering
\includegraphics[width=\textwidth]{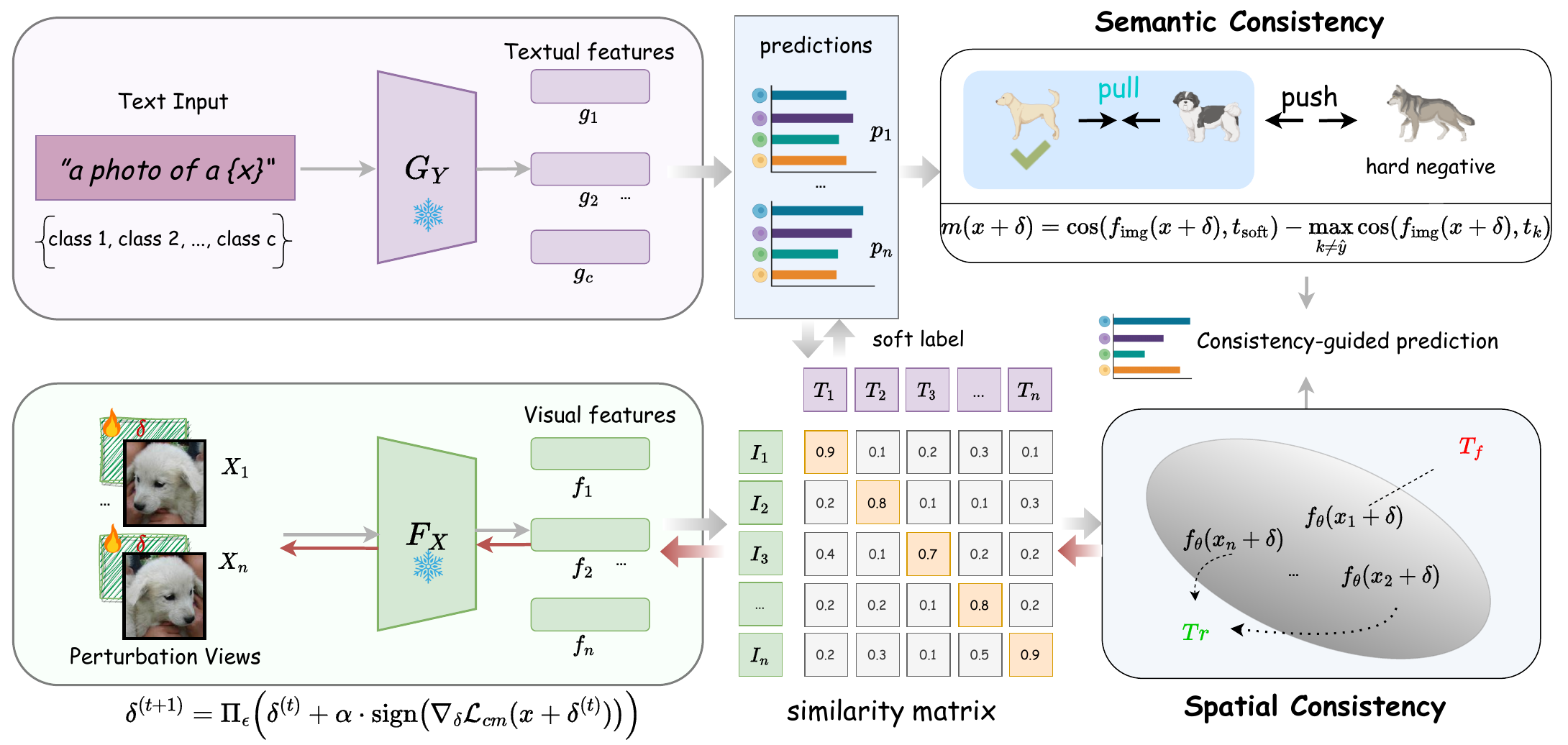}
\caption{
Pipeline of SCC: text and augmented views are encoded into features, multi-view embeddings are aggregated with averaging and combined with a short counterattack warm-up to yield stable soft pseudo-labels, which then guide cross-modal consistency optimization through the corrective perturbation $\delta$.
$T_r$ denotes the correct class embedding (e.g., dog), while $T_f$ is an incorrect class embedding (e.g., wolf). Spatial consistency enforces perturbed views $f_\theta(x_i+\delta)$ to stay close to $T_r$ rather than drift toward $T_f$.
} 
\label{Figure_pipeline}
\end{figure*}

\section{Methodology}
\label{headings}

\subsection{The Findings of Test-time Counterattack}

To motivate our approach, we revisit TTC and identify three vulnerabilities. 
(1) \emph{Semantic drift}: the repaired similarity 
$\cos\!\big(\hat z(x^{\text{cnt}}), \hat t_{y^\star}\big)$, 
with $x^{\text{cnt}}=x^{\text{adv}}+\delta$, often fluctuates and can even shift toward non-target texts under strong attacks (\autoref{analyis-findings}). 
(2) \emph{Hardest-competitor dominance}: misclassifications arise when the repaired embedding aligns closely with the strongest competitor 
$j^\star=\arg\max_{j\neq y^\star}\langle \hat z(x^{\text{cnt}}), \hat t_j\rangle$ 
(\autoref{dog-wolf}). 
(3) \emph{View sensitivity}: across semantics-preserving augmentations $\{v_i\}$ (horizontal flip or low-variance Gaussian noise), 
the repaired logit gaps exhibit high variance, e.g., $\mathrm{Var}_i[\Delta^{(i)}]$ with $\Delta^{(i)}=z^{(i)}_{(1)}-z^{(i)}_{(2)}$, 
indicating inconsistent recovery (\autoref{analyis-findings}).
Together, these expose TTC’s fragility in preserving cross-modal semantics and spatial stability, motivating a principled solution. 
We next formalize \emph{semantic} (1-2) and \emph{spatial fragility} (3), which underpin our SCC framework.

\subsection{The Analysis of Semantic and Spatial Fragility}

Let $\hat f(x)\in\mathbb{R}^d$ denote the $\ell_2$-normalized image embedding and 
$\{\hat t_k\}_{k=1}^K \subset \mathbb{R}^d$ the set of normalized text embeddings.  
The semantic margin of an image $x$ with ground-truth $y^\star$ is
\begin{equation}
    m(x) = \langle \hat f(x), \hat t_{y^\star}\rangle 
    - \max_{j\neq y^\star}\langle \hat f(x), \hat t_j\rangle .
\end{equation}
Under adversarial perturbation $\delta$ with $\|\delta\|_p \le \epsilon$, the margin becomes
\begin{equation}
    m(x+\delta) = \langle \hat f(x+\delta), \hat t_{y^\star}\rangle 
    - \max_{j\neq y^\star}\langle \hat f(x+\delta), \hat t_j\rangle ,
\end{equation}
which often collapses or even turns negative, indicating a shift toward hard negatives.  
This fragility manifests in three forms:

\textbf{Prediction noise.}  
For adversarial inputs $x^{\text{adv}}$, the single-view distribution $\tilde q(y\mid x^{\text{adv}})$ deviates from the ground-truth $p^\star$, introducing
\[
    \text{Bias}=\|\mathbb{E}[\tilde q]-p^\star\|_1,\quad 
    \text{Var}=\sum_{k}\text{Var}[\tilde q_k],
\]
which reduce expected alignment $\mathbb{E}[\langle \hat f(x^{\text{adv}}), t_{y^\star}\rangle]$.

\textbf{Hard-negative alignment.}  
Counterattacks often pull embeddings toward the hardest negative
\[
    j^\star=\arg\max_{j\neq y^\star}\langle \hat f(x^{\text{adv}}), t_j\rangle ,
\]
causing the margin 
$m(x^{\text{adv}})=\langle \hat f(x^{\text{adv}}), t_{y^\star}\rangle-\langle \hat f(x^{\text{adv}}), t_{j^\star}\rangle$
to collapse.

\textbf{View sensitivity.}  
Let $A$ be a distribution of semantics-preserving augmentations.  
Across $N$ sampled views $\{v_i\}$, logits 
$z^{(i)}=\langle \hat f(v_i(x^{\text{adv}})), t_k\rangle$
exhibit high variance $\mathrm{Var}_i[z^{(i)}]$, and the hardest negative $j^\star(i)$ may differ by view.  
Consequently, PGD updates guided by $\nabla_\delta z^{(i)}_{j^\star(i)}$ are inconsistent, yielding large gradient variance $\mathrm{Var}_i[\nabla_\delta \mathcal{L}(z^{(i)})]$ and unstable recovery.  

Together, these effects define \emph{semantic and spatial fragility}, underscoring the difficulty of preserving cross-modal alignment under adversarial perturbations.

\subsection{Mitigating Semantic Fragility via Semantic Consistency}

\paragraph{Cross-modal consistency.}  
Given an adversarial input $x^{\text{adv}}$, the defense applies a counter-perturbation $\delta$ by optimizing a margin objective that encourages alignment with a soft semantic anchor while repelling hard negatives, as shown in \autoref{Figure_pipeline}:
\begin{equation}
\mathcal{L}_{cm}(x^{\text{adv}},\delta) 
= \cos(f_{\text{img}}(x^{\text{adv}}+\delta), t_{\text{soft}}) 
- \max_{k \neq \hat y} \cos(f_{\text{img}}(x^{\text{adv}}+\delta), t_k) ,
\end{equation}
where $\hat y = \arg\max_k \cos(f_{\text{img}}(x^{w}), t_k)$ is pseudo-label predicted from the warm-up embedding $x^{w}$.

\paragraph{Soft prototype construction.}  
To stabilize $t_{\text{soft}}$, we perform a short TTC warm-up (\ref{warm-details}) on $x^{\text{adv}}$ to obtain $x^{w}$, then generate $N$ augmented views $\{v_i(x^{w})\}_{i=1}^N$.  
The view-wise predictions $\{q^{(i)}\}$ are averaged and sharpened with temperature $T<1$:
\begin{equation}
q_k^{\text{sharp}} = \frac{\big(\tfrac{1}{N}\sum_{i=1}^N q^{(i)}_k\big)^{1/T}}
{\sum_j \big(\tfrac{1}{N}\sum_{i=1}^N q^{(i)}_j\big)^{1/T}} ,
\end{equation}
and the soft prototype is defined as
\begin{equation}
t_{\text{soft}} = \sum_k q_k^{\text{sharp}} \, t_k ,
\end{equation}
which acts as the semantic anchor in $\mathcal{L}_{cm}$. The detailed SCC procedure is provided in \autoref{alg1}.

\begin{proposition}[Hard-negative repulsion]
Let $x+\delta$ denote the counter-perturbed input during optimization.  
Optimizing $\mathcal{L}_{cm}$ by PGD ascent increases the semantic margin
\[
m(x+\delta)=\cos(f_{\text{img}}(x+\delta),t_{\text{soft}})
-\max_{k\neq \hat y}\cos(f_{\text{img}}(x+\delta),t_k)
\]
monotonically (up to $\mathcal{O}(\alpha^2)$), thereby preventing drift toward confusable negatives. See proof in Appendix.
\end{proposition}

\paragraph{Iterative counter-attack.}  
Corrective perturbations are computed as
\begin{equation}
\delta^{(t+1)} = \Pi_{\epsilon}\!\left(\delta^{(t)} + \alpha \cdot 
\text{sign}\big(\nabla_\delta \mathcal{L}_{cm}(x+\delta^{(t)})\big)\right),
\end{equation}
where $\Pi_\epsilon$ projects onto the $\ell_p$ ball of radius $\epsilon$ and $\alpha$ is the step size.  
A step-weighted fusion is applied across PGD iterations, where intermediate perturbations $\delta^{(t)}$ are aggregated with weights proportional to their step index, yielding a smoother final correction.

\subsection{Mitigating Spatial Fragility via Spatial Consistency}

\paragraph{Multi-view self-consistency.}  
To stabilize predictions, we aggregate $L$ \emph{augmented views} \citep{sheng2025r} of the same input.  
Let $z^{(i)}$ be the logits of view $i$, then
\[
\bar{z} = \tfrac{1}{L}\sum_{i=1}^L z^{(i)}, \qquad 
\bar{q} = \text{softmax}(\bar{z}), \qquad 
t_{\text{soft}} = \sum_k \bar{q}_k\, t_k .
\]
While each view may yield noisy predictions under adversarial perturbations, their aggregation reduces variance and yields a more reliable semantic anchor.

\begin{proposition}[Variance reduction]
If $\{q^{(i)}\}$ are i.i.d. with covariance $\Sigma$, then $\text{Cov}(\bar{q})=\tfrac{1}{L}\Sigma$, showing variance shrinks as $1/L$ and $t_{\text{soft}}$ becomes more stable.
\end{proposition}

\begin{remark}
Temperature sharpening ($T<1$) further amplifies dominant classes:
$
q_k(T)=\frac{\bar{q}_k^{1/T}}{\sum_j \bar{q}_j^{1/T}},
$
which enlarges semantic margins by suppressing noisy tail classes.
\end{remark}

\paragraph{Confidence weighting.}  
We assign each sample a confidence $w(x)\in[0,1]$ (based on margin or entropy), so that high-confidence predictions dominate optimization, while noisy pseudo-labels are down-weighted. This weighting mitigates error propagation and stabilizes semantic alignment.

\paragraph{Spatial counterattacks.}  
For each input $x$, we first obtain a single counter-perturbation $\delta$ by the TTC inner loop (e.g., PGD-like ascent) under $\|\delta\|_p\le\epsilon$. 
We then form $L$ semantics-preserving views of the corrected image $x+\delta$ via horizontal flip and low-variance Gaussian pixel noise:
\[
\mathcal{V}(x+\delta)=\{\,v_i(x+\delta)\,\}_{i=1}^L,\quad 
v_i(\cdot)=\mathrm{flip}_{\mathrm{h}}^{\mathbf{1}_{i\ \mathrm{odd}}}(\cdot)+\eta_i,\ \ \eta_i\!\sim\!\mathcal{N}(0,(\sigma/255)^2 I).
\]
Let $z^{(i)}$ be the logits of view $i$. We aggregate by averaging logits and then softmax:
\[
\bar z=\tfrac{1}{L}\sum_{i=1}^L z^{(i)},\qquad \hat p=\mathrm{softmax}(\bar z),\qquad \hat y=\arg\max_k \bar z_k.
\]

\begin{table*}[t!]
\centering
\caption{Classification accuracy (\%) on clean images (Acc.) and adversarial images (Rob.) under 10-step PGD attack ($\epsilon_a=1/255$) across 16 datasets. The threat model assumes full access to model weights and gradients. We compare our paradigm against test-time defenses adapted from prior adversarial robustness studies, and include fine-tuned models as references. The last column shows the gains of SCC over the CLIP.}
\label{table-main}
\resizebox{\textwidth}{!}{
\begin{tabular}{c|c|c|cccc|ccccc|c} 
\hline
\multirow{2}{*}{\textbf{Dataset}} & \multirow{2}{*}{Metric} & \multirow{2}{*}{CLIP} & \multicolumn{4}{c|}{\textbf{Adversarial Finetuning}} & \multicolumn{5}{c|}{\textbf{Test-time Defence}}                                   & \multirow{2}{*}{$\Delta$}                                               \\
                                  &                         &                       & CLIP-FT & TeCoA & PMG-AFT & FARE                     & RN    & Anti-adv & HD    & TTC   & SCC(ours)                                      &                                                                         \\ 
\noalign{\kern-\cmidrulewidth}\cmidrule(lr){1-1}\cmidrule(lr){2-2}\cmidrule(lr){3-12}\cmidrule(lr){13-13}
\multirow{2}{*}{CIFAR10}          & Rob.                    & 0.74                  & 3.34    & 33.61 & 40.66   & 19.65                    & 2.01  & 12.39    & 17.22 & 28.75 & {\cellcolor[rgb]{0.969,0.988,1}}\textbf{59.18} & {\cellcolor[rgb]{0.992,0.973,0.973}}\textcolor[rgb]{0.502,0,0}{+58.44}  \\
                                  & Acc.                    & 85.12                 & 84.90   & 64.61 & 70.69   & 74.44                    & 81.18 & 83.52    & 78.23 & 81.18 & {\cellcolor[rgb]{0.969,0.988,1}}82.24          & {\cellcolor[rgb]{0.992,0.973,0.973}}-2.88                               \\
\multirow{2}{*}{CIFAR100}         & Rob.                    & 0.26                  & 0.90    & 18.95 & 22.52   & 11.40                    & 0.67  & 5.73     & 3.86  & 14.31 & {\cellcolor[rgb]{0.969,0.988,1}}\textbf{32.09} & {\cellcolor[rgb]{0.992,0.973,0.973}}\textcolor[rgb]{0.502,0,0}{+31.83}  \\
                                  & Acc.                    & 57.14                 & 59.51   & 35.96 & 40.32   & 46.67                    & 56.34 & 53.95    & 52.86 & 56.34 & {\cellcolor[rgb]{0.969,0.988,1}}55.21          & {\cellcolor[rgb]{0.992,0.973,0.973}}-1.93                               \\
\multirow{2}{*}{STL10}            & Rob.                    & 11.00                 & 12.73   & 70.08 & 73.08   & 59.06                    & 16.23 & 37.42    & 39.02 & 76.70 & {\cellcolor[rgb]{0.969,0.988,1}}\textbf{90.50} & {\cellcolor[rgb]{0.992,0.973,0.973}}\textcolor[rgb]{0.502,0,0}{+79.50}  \\
                                  & Acc.                    & 96.40                 & 94.49   & 87.40 & 88.56   & 91.72                    & 95.85 & 95.45    & 89.50 & 95.85 & {\cellcolor[rgb]{0.969,0.988,1}}95.62          & {\cellcolor[rgb]{0.992,0.973,0.973}}-0.78                               \\
\multirow{2}{*}{ImageNet}         & Rob.                    & 1.15                  & 0.93    & 18.89 & 21.43   & 14.00                    & 1.77  & 8.67     & 6.63  & 38.41 & {\cellcolor[rgb]{0.969,0.988,1}}\textbf{49.77} & {\cellcolor[rgb]{0.992,0.973,0.973}}\textcolor[rgb]{0.502,0,0}{+48.62}  \\
                                  & Acc.                    & 59.69                 & 54.24   & 34.89 & 36.12   & 48.79                    & 59.34 & 54.27    & 54.54 & 49.39 & {\cellcolor[rgb]{0.969,0.988,1}}56.03          & {\cellcolor[rgb]{0.992,0.973,0.973}}-3.66                               \\
\multirow{2}{*}{Caltech101}       & Rob.                    & 14.67                 & 14.21   & 55.51 & 61.08   & 50.74                    & 18.90 & 34.81    & 31.53 & 65.78 & {\cellcolor[rgb]{0.969,0.988,1}}\textbf{77.25} & {\cellcolor[rgb]{0.992,0.973,0.973}}\textcolor[rgb]{0.502,0,0}{+62.58}  \\
                                  & Acc.                    & 85.66                 & 83.63   & 71.68 & 75.45   & 80.95                    & 86.61 & 84.02    & 82.33 & 86.53 & {\cellcolor[rgb]{0.969,0.988,1}}86.44          & {\cellcolor[rgb]{0.992,0.973,0.973}}+0.78                               \\
\multirow{2}{*}{Caltech256}       & Rob.                    & 8.47                  & 6.76    & 43.19 & 45.91   & 38.79                    & 11.33 & 25.36    & 23.48 & 60.11 & {\cellcolor[rgb]{0.969,0.988,1}}\textbf{72.88} & {\cellcolor[rgb]{0.992,0.973,0.973}}\textcolor[rgb]{0.502,0,0}{+64.41}  \\
                                  & Acc.                    & 81.72                 & 78.53   & 61.14 & 62.24   & 73.32                    & 81.25 & 79.38    & 79.12 & 79.66 & {\cellcolor[rgb]{0.969,0.988,1}}81.16          & {\cellcolor[rgb]{0.992,0.973,0.973}}-0.56                               \\
\multirow{2}{*}{OxfordPets}       & Rob.                    & 1.04                  & 2.10    & 38.35 & 41.18   & 31.07                    & 1.86  & 20.42    & 12.04 & 57.87 & {\cellcolor[rgb]{0.969,0.988,1}}\textbf{76.67} & {\cellcolor[rgb]{0.992,0.973,0.973}}\textcolor[rgb]{0.502,0,0}{+75.63}  \\
                                  & Acc.                    & 87.44                 & 84.14   & 62.12 & 65.88   & 79.37                    & 87.41 & 80.62    & 80.91 & 83.35 & {\cellcolor[rgb]{0.969,0.988,1}}86.48          & {\cellcolor[rgb]{0.992,0.973,0.973}}-0.96                               \\
\multirow{2}{*}{Flowers102}       & Rob.                    & 1.14                  & 0.54    & 21.94 & 23.43   & 17.14                    & 1.52  & 7.16     & 7.29  & 39.14 & {\cellcolor[rgb]{0.969,0.988,1}}\textbf{54.59} & {\cellcolor[rgb]{0.992,0.973,0.973}}\textcolor[rgb]{0.502,0,0}{+53.45}  \\
                                  & Acc.                    & 65.46                 & 53.37   & 36.80 & 37.00   & 47.98                    & 64.62 & 62.66    & 58.22 & 64.16 & {\cellcolor[rgb]{0.969,0.988,1}}64.16          & {\cellcolor[rgb]{0.992,0.973,0.973}}-1.30                               \\
\multirow{2}{*}{FGVC-Aircraft}    & Rob.                    & 0.00                  & 0.00    & 2.49  & 2.22    & 1.35                     & 0.00  & 1.27     & 1.26  & 13.77 & {\cellcolor[rgb]{0.969,0.988,1}}\textbf{17.40} & {\cellcolor[rgb]{0.992,0.973,0.973}}\textcolor[rgb]{0.502,0,0}{+17.40}  \\
                                  & Acc.                    & 20.10                 & 14.04   & 5.31  & 5.55    & 10.86                    & 19.25 & 15.88    & 16.36 & 18.00 & {\cellcolor[rgb]{0.969,0.988,1}}17.61          & {\cellcolor[rgb]{0.992,0.973,0.973}}-2.49                               \\
\multirow{2}{*}{StanfordCars}     & Rob.                    & 0.02                  & 0.06    & 8.76  & 11.65   & 6.75                     & 0.16  & 4.40     & 2.71  & 33.01 & {\cellcolor[rgb]{0.969,0.988,1}}\textbf{43.24} & {\cellcolor[rgb]{0.992,0.973,0.973}}\textcolor[rgb]{0.502,0,0}{+43.22}  \\
                                  & Acc.                    & 52.02                 & 42.11   & 20.91 & 25.44   & 38.68                    & 52.14 & 36.21    & 44.28 & 48.16 & {\cellcolor[rgb]{0.969,0.988,1}}51.19          & {\cellcolor[rgb]{0.992,0.973,0.973}}-0.83                               \\
\multirow{2}{*}{SUN397}           & Rob.                    & 1.14                  & 0.94    & 19.39 & 22.58   & 14.91                    & 1.72  & 8.05     & 6.40  & 41.52 & {\cellcolor[rgb]{0.969,0.988,1}}\textbf{53.27} & {\cellcolor[rgb]{0.992,0.973,0.973}}\textcolor[rgb]{0.502,0,0}{+52.13}  \\
                                  & Acc.                    & 58.50                 & 55.73   & 36.69 & 37.98   & 52.42                    & 59.69 & 56.00    & 53.17 & 55.13 & {\cellcolor[rgb]{0.969,0.988,1}}58.25          & {\cellcolor[rgb]{0.992,0.973,0.973}}-0.25                               \\
\multirow{2}{*}{Country211}       & Rob.                    & 0.04                  & 0.03    & 1.78  & 2.12    & 0.85                     & 0.06  & 0.67     & 0.47  & 7.09  & {\cellcolor[rgb]{0.969,0.988,1}}\textbf{9.41}  & {\cellcolor[rgb]{0.992,0.973,0.973}}\textcolor[rgb]{0.502,0,0}{+9.37}   \\
                                  & Acc.                    & 15.25                 & 12.07   & 4.75  & 4.64    & 9.26                     & 14.80 & 11.58    & 11.72 & 13.08 & {\cellcolor[rgb]{0.969,0.988,1}}13.36          & {\cellcolor[rgb]{0.992,0.973,0.973}}-1.89                               \\
\multirow{2}{*}{Food101}          & Rob.                    & 0.70                  & 0.42    & 13.90 & 18.57   & 11.65                    & 1.20  & 13.12    & 8.03  & 57.84 & {\cellcolor[rgb]{0.969,0.988,1}}\textbf{65.39} & {\cellcolor[rgb]{0.992,0.973,0.973}}\textcolor[rgb]{0.502,0,0}{+64.69}  \\
                                  & Acc.                    & 83.88                 & 64.86   & 29.98 & 36.61   & 55.31                    & 83.44 & 75.81    & 80.30 & 82.18 & {\cellcolor[rgb]{0.969,0.988,1}}82.13          & {\cellcolor[rgb]{0.992,0.973,0.973}}-1.75                               \\
\multirow{2}{*}{EuroSAT}          & Rob.                    & 0.03                  & 0.04    & 11.96 & 12.60   & 10.67                    & 0.15  & 2.15     & 4.57  & 12.19 & {\cellcolor[rgb]{0.969,0.988,1}}\textbf{20.64} & {\cellcolor[rgb]{0.992,0.973,0.973}}\textcolor[rgb]{0.502,0,0}{+20.61}  \\
                                  & Acc.                    & 42.59                 & 27.64   & 16.58 & 18.53   & 21.88                    & 53.24 & 36.78    & 39.08 & 53.24 & {\cellcolor[rgb]{0.969,0.988,1}}41.69          & {\cellcolor[rgb]{0.992,0.973,0.973}}-0.90                               \\
\multirow{2}{*}{DTD}              & Rob.                    & 2.98                  & 2.39    & 17.61 & 14.95   & 15.64                    & 3.71  & 5.62     & 11.63 & 27.32 & {\cellcolor[rgb]{0.969,0.988,1}}\textbf{34.57} & {\cellcolor[rgb]{0.992,0.973,0.973}}\textcolor[rgb]{0.502,0,0}{+31.59}  \\
                                  & Acc.                    & 40.64                 & 36.49   & 25.16 & 21.76   & 32.07                    & 37.96 & 38.92    & 34.89 & 36.98 & {\cellcolor[rgb]{0.969,0.988,1}}37.34          & {\cellcolor[rgb]{0.992,0.973,0.973}}-3.30                               \\
\multirow{2}{*}{PCAM}             & Rob.                    & 0.08                  & 1.11    & 48.24 & 46.18   & 16.23                    & 0.41  & 4.97     & 44.74 & 52.85 & {\cellcolor[rgb]{0.969,0.988,1}}\textbf{69.99} & {\cellcolor[rgb]{0.992,0.973,0.973}}\textcolor[rgb]{0.502,0,0}{+69.91}  \\
                                  & Acc.                    & 52.02                 & 47.21   & 49.96 & 50.03   & 52.54                    & 52.73 & 52.49    & 50.38 & 52.73 & {\cellcolor[rgb]{0.969,0.988,1}}54.41          & {\cellcolor[rgb]{0.992,0.973,0.973}}+2.39                               \\ 
\hline\hline
\multirow{2}{*}{Avg.}             & Rob.                    & 2.70                  & 2.91    & 26.54 & 28.76   & 20.00                    & 3.86  & 12.01    & 13.81 & 39.17 & {\cellcolor[rgb]{0.969,0.988,1}}\textbf{51.68} & {\cellcolor[rgb]{0.992,0.973,0.973}}\textcolor[rgb]{0.502,0,0}{+48.98}  \\
                                  & Acc.                    & 61.51                 & 55.80   & 40.25 & 42.30   & 51.02                    & 61.61 & 57.35    & 56.62 & 59.75 & {\cellcolor[rgb]{0.969,0.988,1}}60.21          & {\cellcolor[rgb]{0.992,0.973,0.973}}-1.30                               \\
\hline
\end{tabular}}
\end{table*}

\begin{remark}[Optimization coupling]
Unlike pure test-time ensembling, all augmented views share a common corrective perturbation $\delta$, which is optimized jointly in the TTC loop. This coupling enforces spatial consistency while repairing adversarial effects.
\end{remark}

\begin{proposition}[Suppression of spurious negatives]
For averaged logits, 
\[
\max_{j\neq y^\star}\bar{z}_j \;\leq\; \tfrac{1}{L}\sum_i \max_{j\neq y^\star}z^{(i)}_j,
\]
so aggregation suppresses view-dependent hardest negatives and stabilizes TTC updates. See proof in Appendix.
\end{proposition}

\paragraph{Objective.}  
The SCC optimization couples cross-modal semantics with spatial stability:
\begin{equation}
\max_{\|\delta\|\leq\epsilon} \;\; \lambda_{cm} \mathcal{L}_{cm}(x,\delta) 
+  \|f_{\text{img}}(x+\delta)-f_{\text{img}}(x)\|_2^2,
\end{equation}
where the second term follows TTC in promoting feature deviation to escape pseudo-stability.  
\begin{remark}
This unifies semantic alignment and spatial consistency into a defense objective.
\end{remark}

\section{Experiments}
\label{others}

\subsection{Experimental Setup}

\textbf{Datasets and Baselines:}
Building on prior studies of CLIP’s adversarial robustness \citep{mao2022understanding,xing2025clip}, we evaluate on 16 public datasets spanning diverse visual domains: generic object recognition (CIFAR10 \citep{krizhevsky2012imagenet}, CIFAR100 \citep{krizhevsky2012imagenet}, STL10 \citep{coates2011analysis}, ImageNet \citep{deng2009imagenet}, Caltech101 \citep{fei2006one}, Caltech256 \citep{griffin2008learning}), fine-grained recognition (OxfordPets \citep{parkhi2012cats}, Flowers102 \citep{nilsback2008automated}, Food101 \citep{bossard2014food}, StanfordCars \citep{krause20133d}), scene recognition (SUN397 \citep{xiao2010sun}, Country211 \citep{radford2021learning}), and specialized domains (FGVCAircraft \citep{maji2013fine}, EuroSAT \citep{helber2019eurosat}, DTD \citep{cimpoi2014describing}, PCAM \citep{bejnordi2017diagnostic}). Comprehensive evaluation further includes experiments on 6 medical datasets such as BUSI \citep{al2020dataset}, BTMRI \citep{koleilat2025biomedcoop}, CHMNIST \citep{kather2016multi}, COVID-19 \citep{tahir2021covid}, DermaMNIST \citep{codella2019skin}, and KneeXray \citep{chen2018knee}.

We implemented several baselines for comparison. Test-time defenses include Test-time Counterattack (TTC) \citep{xing2025clip}, following the original setup, Anti-Adversarial \citep{alfarra2022combating} (adapted to CLIP by maximizing image–text similarity), Hedging Defense (HD) \citep{wu2021attacking} (minimizing cross-entropy across all classes), and RN, which perturbs inputs with random noise of the same strength as $\epsilon$ \citep{xing2025clip}.
As reference, we evaluated adversarial fine-tuning methods—TeCoA \citep{mao2022understanding}, PMG-AFT \citep{wang2024pre}, FARE \citep{schlarmann2024robust}—and a clean fine-tuned CLIP (CLIP-FT) on TinyImageNet, using 2-step PGD ($\alpha=1/255$, $\epsilon_a=1/255$) and learning rate $5\!\times\!10^{-5}$, then transferring the models to 16 downstream datasets.

\begin{table*}
\centering
\setlength{\extrarowheight}{0pt}
\addtolength{\extrarowheight}{\aboverulesep}
\addtolength{\extrarowheight}{\belowrulesep}
\setlength{\aboverulesep}{0pt}
\setlength{\belowrulesep}{0pt}
\caption{Adversarial (Rob.) and clean (Acc.) accuracy (\%) on 16 datasets under PGD-10 ($\epsilon_a=4/255$). Superscripts denote fine-tuning budgets. The last row shows gains over CLIP.}
\label{table-4/255}
\resizebox{\textwidth}{!}{
\begin{tabular}{c|cccccccc|cccc|cc} 
\toprule
(\%) & CLIP  & CLIP-FT & TeCoA$^1$ & TeCoA$^4$ & PMG-AFT$^1$ & PMG-AFT$^4$ & FARE$^1$ & FARE$^4$ & RN    & Anti-adv & HD    & TTC   & {\cellcolor[rgb]{0.969,0.988,1}}SCC(ours)      & {\cellcolor[rgb]{0.992,0.973,0.973}}$\Delta$         \\ 
\hhline{-|--------|----|--}
Rob. & 0.09  & 0.96    & 6.51      & 10.03     & 7.03        & 10.70       & 1.50     & 3.67     & 0.06  & 0.53     & 1.19  & 20.63 & {\cellcolor[rgb]{0.969,0.988,1}}\textbf{27.88} & {\cellcolor[rgb]{0.992,0.973,0.973}}+\textcolor[rgb]{0.502,0,0}{27.79}  \\
Acc. & 61.51 & 55.80   & 40.25     & 35.57     & 42.30       & 37.58       & 51.02    & 46.17    & 61.61 & 57.32    & 56.62 & 55.99 & {\cellcolor[rgb]{0.969,0.988,1}}\textbf{60.42} & {\cellcolor[rgb]{0.992,0.973,0.973}}-1.09            \\
\bottomrule
\end{tabular}}
\end{table*}

\textbf{Implementation:} 
We adopt CLIP ViT-B/32 as the backbone \citep{radford2021learning} and BioMedCLIP \citep{zhang2025multimodal} for medical tasks, using the handcrafted prompt templates from CLIP. Counterattack budget are set to $\epsilon=4/255$, and 2 steps \citep{xing2025clip}. For the semantic consistency, $\lambda_{cm}=4$ and temperature $T=0.5$ (selected via grid search). For the spatial consistency, we use $L=2$ augmented views with noise $\sigma=6$ (tuned by search). 
We evaluate against white-box and adaptive attacks, including $PGD\text{-}\ell_\infty$ and CW \citep{xing2025clip}. 
By default, we report top-1 accuracy on both clean and adversarial examples. Counterattack parameters follow \citep{xing2025clip}. The batch size is set to 256. We conducted all experiments on NVIDIA H20 GPUs.

\subsection{Main Results}
We evaluate robustness under an attack budget of $\epsilon_{a}=1/255$, following prior CLIP robustness studies \citep{xing2025clip}. All baselines are tested on 16 datasets with 10-step PGD attacks, assuming full access to model weights and gradients but no access to test-time operations. As shown in \autoref{table-main}, adversarially fine-tuned models (TeCoA, PMG-AFT, FARE, CLIP-FT) suffer from severe overfitting: while robust accuracy improves on training-like datasets, clean accuracy drops significantly across downstream tasks. 
Among test-time defenses, Anti-Adversarial and HD yield only marginal gains, while RN fails to provide robustness even with perturbations much larger than $\epsilon_a$. TTC delivers noticeable gains but falls significantly short of SCC.
In contrast, our SCC achieves consistent improvements: the average robust accuracy rises from 2.70\% (CLIP) and 39.17\% (TTC) to 51.68\%, a substantial gain of +48.98\% over vanilla CLIP and +12.51\% over TTC, with clean accuracy only slightly reduced ($-$1.30\%). These results highlight SCC as an test-time defense that delivers strong and stable adversarial robustness without sacrificing clean performance.

We further evaluate robustness under a stronger attack budget $\epsilon_{a}=4/255$. 
For the stronger-budget setting, we increase counterattack iterations to $5$ while keeping all other hyperparameters fixed; adversarial fine-tuning baselines are trained with the same perturbation budget.
As shown in \autoref{table-4/255} and \ref{eps4}, robust accuracy of all models drops significantly under stronger attacks. Anti-Adversarial and HD almost lose robustness in this setting, while TTC provides moderate protection but suffers from high variance across datasets. In contrast, our SCC achieves stable improvements: average robust accuracy rises to 27.88\%, outperforming TTC by +7.25\% and vanilla CLIP by +27.79\%, with only a negligible clean accuracy drop (–1.09\%). These results demonstrate that SCC remains effective even under high-budget adversarial perturbations, highlighting its robustness and generalization.
Per-dataset results are provided in the Appendix.
We further evaluate SCC under CW attacks \citep{carlini2017towards}, with results deferred to the Appendix due to space limits (\ref{sec_cw}).

Adversarial robustness in the medical domain is particularly challenging: as shown in \autoref{table-medical}, BioMedCLIP nearly collapses under $\epsilon_{a}=1/255$ attacks, with average adversarial accuracy close to $0\%$ \citep{koleilat2025biomedcoop}. TTC alleviates this issue by introducing counterattacks, improving robustness to 10.62\% on average. Our SCC further restores robustness substantially, reaching 34.00\% on BioMedCLIP (a +23.38\% improvement over TTC) while maintaining clean accuracy (44.63\%). On CLIP, SCC also consistently outperforms TTC across six medical datasets, improving robustness by +5.94\% on average. These results demonstrate that SCC not only generalizes to domain-specific models like BioMedCLIP but also provides a plug-and-play defense that stabilizes zero-shot medical prediction under adversarial perturbations. 

\begin{table*}
\centering
\setlength{\extrarowheight}{0pt}
\addtolength{\extrarowheight}{\aboverulesep}
\addtolength{\extrarowheight}{\belowrulesep}
\setlength{\aboverulesep}{0pt}
\setlength{\belowrulesep}{0pt}
\caption{CLIP and BioMedCLIP Robustness on Medical Benchmarks ($\epsilon_{a}=1/255$).}
\resizebox{\textwidth}{!}{
\begin{tabular}{c|ccccccccccccccc} 
\toprule
Backbone                    &            & \multicolumn{2}{c}{BUSI}                               & \multicolumn{2}{c}{BTMRI}                              & \multicolumn{2}{c}{CHMNIST}                            & \multicolumn{2}{c}{COVID\_19}                          & \multicolumn{2}{c}{DermaMNIST}                         & \multicolumn{2}{c}{KneeXray}                           & \multicolumn{2}{c}{Avg.}                                \\ 
\hhline{----------------}
\multirow{4}{*}{CLIP}       &            & {\cellcolor[rgb]{0.969,0.988,1}}Rob.           & Acc.  & {\cellcolor[rgb]{0.969,0.988,1}}Rob.           & Acc.  & {\cellcolor[rgb]{0.969,0.988,1}}Rob.           & Acc.  & {\cellcolor[rgb]{0.969,0.988,1}}Rob.           & Acc.  & {\cellcolor[rgb]{0.969,0.988,1}}Rob.           & Acc.  & {\cellcolor[rgb]{0.969,0.988,1}}Rob.           & Acc.  & {\cellcolor[rgb]{0.969,0.988,1}}Rob.           & Acc.   \\
                            & CLIP       & {\cellcolor[rgb]{0.969,0.988,1}}0.00           & 47.18 & {\cellcolor[rgb]{0.969,0.988,1}}0.00           & 25.60 & {\cellcolor[rgb]{0.969,0.988,1}}0.00           & 21.32 & {\cellcolor[rgb]{0.969,0.988,1}}0.13           & 6.39  & {\cellcolor[rgb]{0.969,0.988,1}}0.02           & 19.76 & {\cellcolor[rgb]{0.969,0.988,1}}0.00           & 13.74 & {\cellcolor[rgb]{0.969,0.988,1}}0.02           & 22.33  \\
                            & TTC        & {\cellcolor[rgb]{0.969,0.988,1}}11.67          & 42.05 & {\cellcolor[rgb]{0.969,0.988,1}}8.93           & 27.84 & {\cellcolor[rgb]{0.969,0.988,1}}2.20           & 19.64 & {\cellcolor[rgb]{0.969,0.988,1}}\textbf{7.51}  & 9.11  & {\cellcolor[rgb]{0.969,0.988,1}}6.28           & 20.68 & {\cellcolor[rgb]{0.969,0.988,1}}7.68           & 14.32 & {\cellcolor[rgb]{0.969,0.988,1}}7.38           & 22.27  \\
                            & SCC        & {\cellcolor[rgb]{0.969,0.988,1}}\textbf{23.85} & 42.31 & {\cellcolor[rgb]{0.969,0.988,1}}\textbf{16.19} & 27.78 & {\cellcolor[rgb]{0.969,0.988,1}}\textbf{9.12}  & 17.26 & {\cellcolor[rgb]{0.969,0.988,1}}7.30           & 7.05  & {\cellcolor[rgb]{0.969,0.988,1}}\textbf{12.40} & 20.48 & {\cellcolor[rgb]{0.969,0.988,1}}\textbf{11.08} & 13.12 & {\cellcolor[rgb]{0.969,0.988,1}}\textbf{13.32} & 21.33  \\ 
\hhline{~---------------}\cmidrule(lr){1-1}
\multirow{3}{*}{BioMedCLIP} & BioMedCLIP & {\cellcolor[rgb]{0.969,0.988,1}}0.00           & 40.38 & {\cellcolor[rgb]{0.969,0.988,1}}0.49           & 60.33 & {\cellcolor[rgb]{0.969,0.988,1}}0.00           & 32.62 & {\cellcolor[rgb]{0.969,0.988,1}}0.02           & 72.53 & {\cellcolor[rgb]{0.969,0.988,1}}0.00           & 35.62 & {\cellcolor[rgb]{0.969,0.988,1}}0.00           & 27.92 & {\cellcolor[rgb]{0.969,0.988,1}}0.08           & 44.90  \\
                            & TTC        & {\cellcolor[rgb]{0.969,0.988,1}}7.95           & 37.05 & {\cellcolor[rgb]{0.969,0.988,1}}22.20          & 53.08 & {\cellcolor[rgb]{0.969,0.988,1}}2.80           & 29.62 & {\cellcolor[rgb]{0.969,0.988,1}}18.36          & 57.20 & {\cellcolor[rgb]{0.969,0.988,1}}4.91           & 24.58 & {\cellcolor[rgb]{0.969,0.988,1}}7.51           & 34.10 & {\cellcolor[rgb]{0.969,0.988,1}}10.62          & 39.27  \\
                            & SCC        & {\cellcolor[rgb]{0.969,0.988,1}}\textbf{31.92} & 40.26 & {\cellcolor[rgb]{0.969,0.988,1}}\textbf{48.93} & 59.72 & {\cellcolor[rgb]{0.969,0.988,1}}\textbf{16.56} & 31.24 & {\cellcolor[rgb]{0.969,0.988,1}}\textbf{57.58} & 68.95 & {\cellcolor[rgb]{0.969,0.988,1}}\textbf{20.64} & 32.51 & {\cellcolor[rgb]{0.969,0.988,1}}\textbf{28.35} & 35.07 & {\cellcolor[rgb]{0.969,0.988,1}}\textbf{34.00} & 44.63  \\
\bottomrule
\end{tabular}}
\label{table-medical}
\end{table*}

\subsection{Ablation Studies}

\begin{figure*}[t]
\centering
\includegraphics[width=\textwidth]{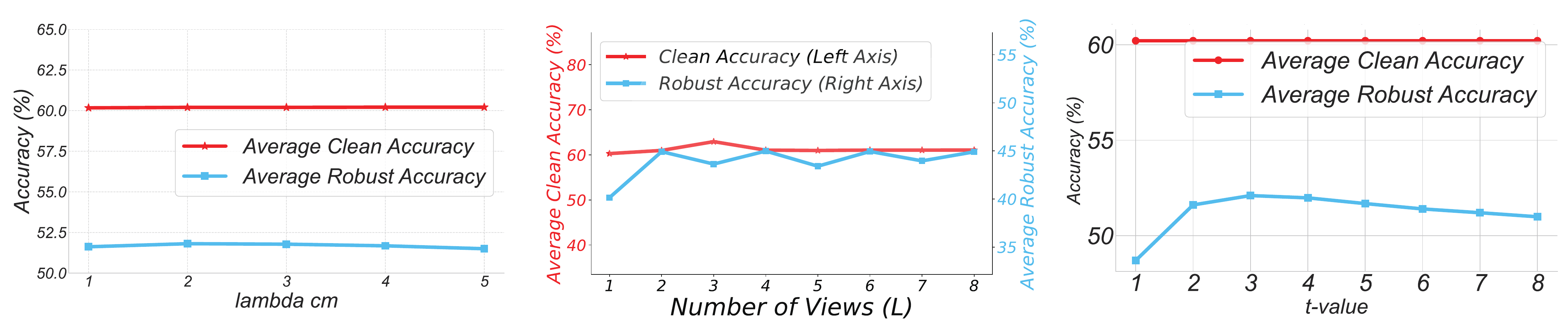}
\caption{
 Sensitivity of SCC to $\lambda_{cm}$, number of views $L$, and effect of the temperature $t$ in soft-label sharpening (The $t$-axis in the plot is scaled by $\times 10$). A moderate $t$ yields the best trade-off.
} 
\label{Figure_zhexian1}
\end{figure*}

\begin{figure*}[t]
\centering
\includegraphics[width=\textwidth]{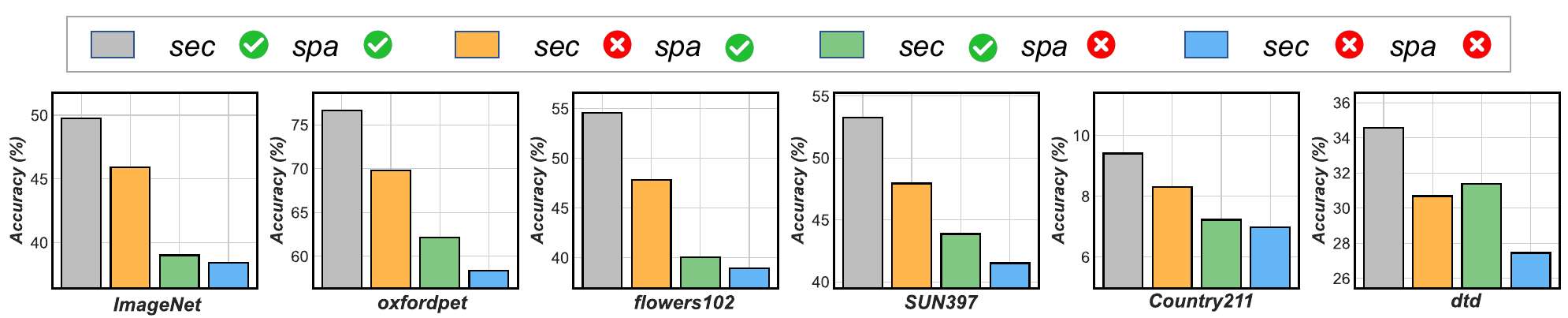}
\caption{
Ablation results of semantic consistency (sec) and spatial consistency (spa) across datasets. Removing either component degrades performance, while combining both yields the best robustness and accuracy.
} 
\label{Figure_ablation}
\end{figure*}

\textbf{Effect of self-calibrated consistency:}
\autoref{Figure_ablation} and \autoref{tab:ablation} ablate SCC’s two modules on individual datasets and averaged over 16 datasets. As shown in \autoref{Figure_ablation}, retaining both semantic (Sec) and spatial (Spa) consistency yields the best performance across all datasets, while removing either leads to sharp drops, and removing both results in the lowest accuracy. Concretely, with only semantic consistency, robustness is 39.76\% (clean 60.28\%); with only spatial consistency, 48.01\% (clean 59.76\%). Combining both raises robustness to 51.68\% (clean 60.21\%), and while removing both modules drops the performance to -12.5\% robustness and -0.48\% clean accuracy. These results confirm the complementarity of the two modules: each provides modest gains alone, but together they deliver substantial robustness improvements while maintaining clean accuracy.

\textbf{Analysis of hyperparameter sensitivity.} 
We conducted grid searches over the key hyperparameters of SCC. As shown in Figure 5, the cross-modal regularization weight $\lambda_{cm}\in[1,5]$ has little effect on clean accuracy and only mild impact on robustness, with a small peak around the mid-range; we adopt $\lambda_{cm}=4$ for stability. The number of views $L$ strongly influences robustness, which increases sharply from $L=1$ to $L=2$–$4$ before saturating; we set $L=2$ for a balance of accuracy and efficiency. The temperature $T$ used in soft-label sharpening (\autoref{Figure_zhexian1}) also affects robustness, with $T=0.5$ yielding the best trade-off. Finally, the noise scale $\sigma$ (\autoref{Figure_analsis-vis-2}) steadily boosts robustness until saturation, at the cost of a slight clean accuracy drop; we adopt $\sigma=6$. Overall, SCC is not overly sensitive to hyperparameter choices, and the selected defaults yield strong robustness gains with minimal accuracy loss.

\vspace{-1em}


\subsection{Visualization and Efficiency Analysis.}
\autoref{Figure_vis} illustrates the effect of SCC on CIFAR-10 under adversarial attacks.
In panel (a), the distribution of maximum soft-label probabilities shows that, compared to the adversarial case (red), SCC (blue) shifts the distribution closer to clean samples (green), indicating better calibration and reduced over-confidence. Panels (b) and (c) compare confusion matrices: without SCC (b), adversarial perturbations induce widespread misclassifications, whereas with SCC (c), diagonal dominance is largely restored, confirming improved accuracy and stability across categories.
In terms of efficiency, \autoref{tab:time} shows that, unlike R-TPT which requires many view transformations, both TTC and SCC achieve much lower inference overhead. Notably, SCC incurs only an additional 0.0005s per image compared to TTC, yet delivers a +7.2\% gain in robustness. This demonstrates SCC’s clear superiority in achieving a favorable trade-off between robustness and efficiency.

\begin{table*}[ht!]
\centering
\begin{minipage}[t]{0.48\textwidth}
    \centering
    \resizebox{\linewidth}{!}{%
    \begin{tabular}{@{}lc|cc@{}}
    \toprule
    \textbf{Method} & \textbf{Stage} & \textbf{Time} & \textbf{Rob.} \\
    \midrule
    R-TPT (64 views) & Test time & 0.37s/img  & 32.8 \\
    TTC              & Test time & 0.012s/img & 27.4 \\
    SCC (ours)       & Test time & 0.0125s/img & \textbf{34.6} \\
    \bottomrule
    \end{tabular}}
    \caption{Running time and adversarial accuracies (\%) of methods against adversarial attack on DTD dataset.}
    \label{tab:time}
\end{minipage}
\hfill
\begin{minipage}[t]{0.48\textwidth}
    \centering
    \resizebox{\linewidth}{!}{%
    \begin{tabular}{cccc} 
    \toprule
    Semantic Consistency & Spatial Consistency & Rob. & Acc. \\ 
    \midrule
                         &                     & 39.18 & 59.73 \\
    \cmark               &                     & 39.76 & 60.28 \\
                         & \cmark              & 48.01 & 59.76 \\
    \cmark               & \cmark              & \textbf{51.68} & 60.21 \\
    \bottomrule
    \end{tabular}}
    \caption{Ablation of semantic and spatial consistency across 16 datasets.}
    \label{tab:ablation}
\end{minipage}
\end{table*}

\section{Conclusion}
In this paper, we presented SCC, a test-time defense that strengthens the adversarial robustness of vision–language models in the zero-shot setting. SCC unifies two complementary components: 
semantic consistency, which resists cross-modal drift by repelling hard negatives, 
and spatial consistency, which stabilizes predictions through multi-view augmentation and correction. 
Extensive experiments across 22 benchmarks, including the domain-specific BioMedCLIP model, show that SCC yields consistent gains in robustness with minimal loss of clean accuracy. 
Our results demonstrate that SCC offers a simple and effective way to enhance the reliability of VLMs across both general-purpose and safety-critical domains.

\section*{Ethics Statement}

This work uses only publicly available datasets without personal or sensitive information.  
By improving adversarial robustness of vision--language models, SCC aims to enhance reliability in both general and medical applications.

\section*{Reproducibility Statement}

We provide implementation details in Experments and Appendix, including algorithmic descriptions, and hyperparameters.
All experiments are conducted on publicly available datasets, and our code with scripts for reproducing results will be released upon publication.

\bibliography{iclr2026_conference}
\bibliographystyle{iclr2026_conference}

\appendix
\section{Appendix}

\begin{figure*}[ht!]
\centering
\includegraphics[width=\textwidth]{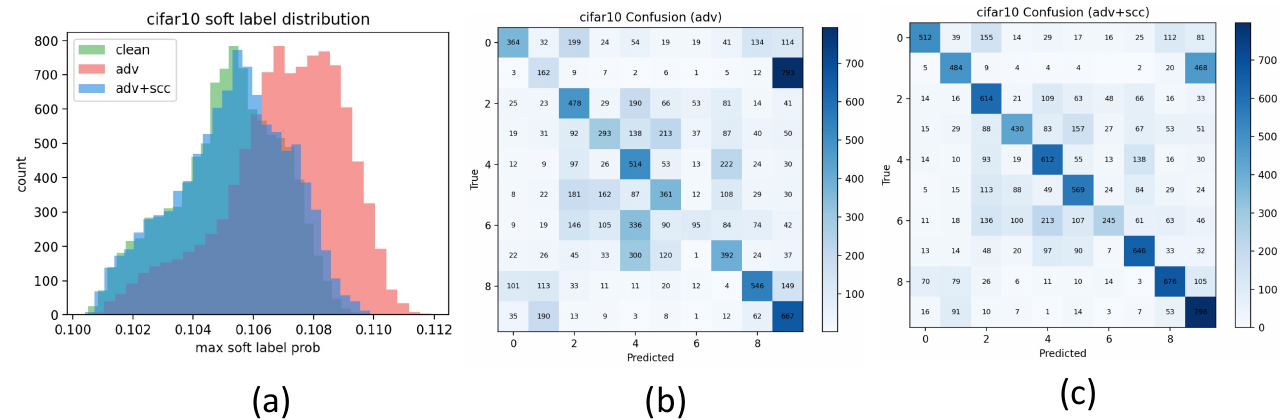}
\caption{
 (a) Distribution of maximum soft label probabilities for clean, adversarial, and adversarial+SCC samples on CIFAR-10. SCC shifts the distribution toward clean probability. (b) Confusion matrix for adversarial samples, showing increased misclassification. (c) Confusion matrix for adversarial samples with SCC, demonstrating improved classification accuracy and reduced confusion
} 
\label{Figure_vis}
\end{figure*}

\subsection{Implementation Detail}
\label{warm-details}
For the counterattack analysis (\autoref{analyis-findings}), we compare three settings. In the \emph{multi-view} case, two augmented views (horizontal flip) are used to construct predictions. The \emph{single-view} case reduces this to one view, removing variance reduction. For the \emph{semantic perturbation} case, we randomly insert additional words into the text prompts, which distorts cross-modal alignment. Results in \autoref{analyis-findings} show that reducing to a single view significantly decreases robustness, and adding random semantic perturbations further degrades performance.

We perform a short TTC warm-up on each adversarial input $x^{\text{adv}}$ using PGD-like steps ($\epsilon=4/255$, $\alpha=1/255$), optimizing only the feature-deviation term to avoid label bias. Instead of early stopping, perturbations from all steps are fused by a $\tau$-threshold weighting scheme, 
yielding a stabilized initialization $x^w$. On $x^w$, $N$ lightweight augmented views \cite{sheng2025r} (flip + Gaussian noise with $\sigma=6/255$) are generated, their logits averaged before softmax, and the sharpened distribution ($T=0.5$) used to construct the soft prototype $t_{\text{soft}}$, which serves as the semantic anchor in subsequent optimization.

\subsection{Performance of SCC guided by clean image predictions}
\autoref{fig_clean} reports results when SCC is guided by predictions on clean images. Under this setting, SCC achieves an average robust accuracy of 53.62\% with clean accuracy of 61.78\%, showing a +50.92\% improvement over vanilla CLIP. However, this requires access to clean-image predictions at inference, which is impractical in real-world deployment. By contrast, our pseudo-labeling strategy achieves 51.68\% robust accuracy, closely approaching the clean-prediction upper bound while remaining label-free and deployable.

\subsection{Robustness under CW attacks}
\label{sec_cw}
Following prior CLIP robustness studies, we evaluate under a 10-step CW attack \citep{carlini2017towards} with budget $\epsilon_{a}=1/255$ across 16 datasets (white-box access to weights/gradients). As shown in \autoref{table-cw}, SCC attains the highest average robust accuracy, 49.42\%, improving over vanilla CLIP by +45.88\% and over the strongest test-time baseline (TTC) by large margins, while keeping clean accuracy essentially unchanged (60.21\%, -1.30\%). RN and TTE preserve clean accuracy (they do not counter-perturb inputs) but offer limited or unstable robustness. Anti-Adversarial and HD, which optimize targeted perturbations, yield low robust accuracy and further reduce clean performance. Adversarially fine-tuned models increase robustness on some datasets but at a substantial clean-accuracy cost. Overall, SCC consistently delivers the best robustness–accuracy trade-off under CW, indicating that inference-time self-calibrated consistency generalizes beyond PGD to stronger optimization-based attacks.

\subsection{Analysis of Robustness (under $\epsilon_{a}=4/255$)}
\label{eps4}
\autoref{table_4/255-2} summarizes robustness under a stronger 10-step PGD attack with budget $\epsilon_{a}=4/255$ across 16 datasets. We observe that Anti-Adversarial and HD almost collapse under this setting, offering negligible robustness. RN maintain high clean accuracy, as they do not introduce counter-perturbations, but RN provides no robustness and TTE exhibits highly unstable gains, as reflected by large standard deviations across runs. By contrast, SCC consistently improves robustness across all datasets, achieving an average robust accuracy of 27.88\%, a gain of +27.79\% over vanilla CLIP, while keeping clean accuracy largely intact (60.42\%, $-1.09$\%). 
To further strengthen counterattacks under this high-budget regime, we increase the iteration number to $N=5$ for TTC. Although this slightly reduces clean accuracy by 5.52 points compared to CLIP, the substantial robustness gains justify the trade-off. Overall, these results confirm that SCC maintains stable and significant robustness improvements even under stronger adversarial budgets.

\subsection{Effects of Other Hyperparameters}
\begin{figure*}[t]
\centering
\includegraphics[width=\textwidth]{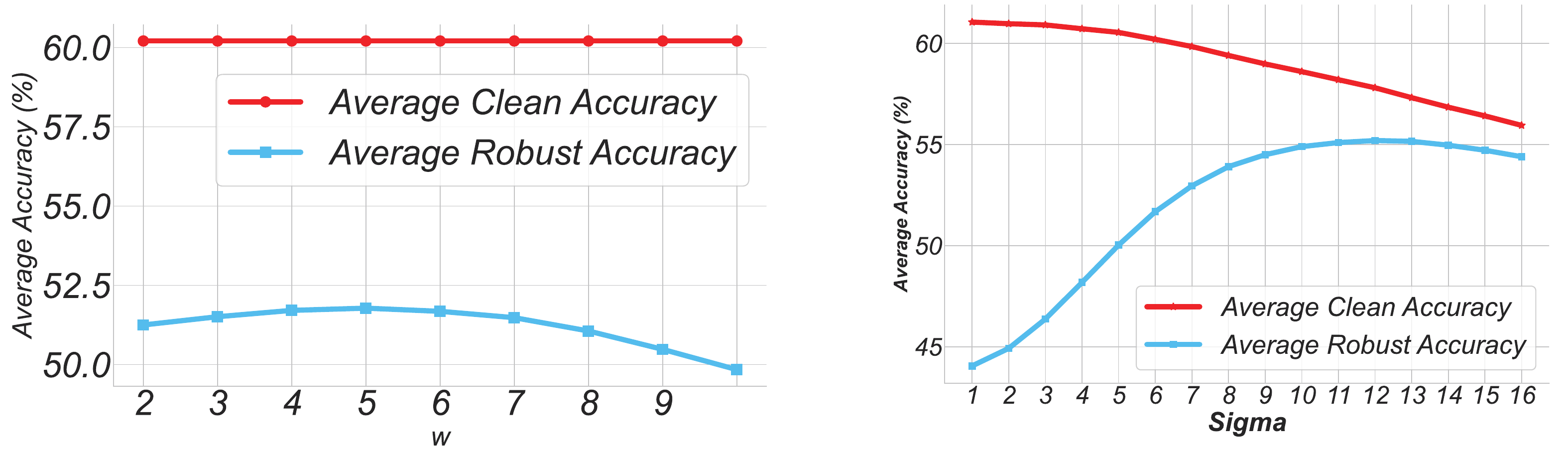}
\caption{
 Effect of the warm-up step number $w$ in short TTC: a moderate number yields the best robustness, while clean accuracy is unaffected. Effect of the Gaussian noise scale $\sigma$ (Sigma). Robustness improves with more views and larger $\sigma$, while clean accuracy drops.
} 
\label{Figure_analsis-vis-2}
\end{figure*}

We further analyze the impact of additional hyperparameters on SCC. As shown in \autoref{Figure_analsis-vis-2}, increasing the warm-up steps $w$ used for generating pseudo-labels leads to stable clean accuracy but only marginal gains in robustness, which peaks around $w=5$ before declining. This indicates that a small number of warm-up iterations is sufficient to stabilize pseudo-label quality without introducing excessive counter-perturbations. Therefore, we set $w=5$. 
On the other hand, the noise scale $\sigma$ for multi-view augmentation plays a more critical role. Larger $\sigma$ significantly boosts robustness by enhancing view diversity, while clean accuracy decreases gradually as perturbations grow stronger. Overall, SCC exhibits stable behavior across a wide range of hyperparameters, with robustness consistently improving under larger $\sigma$ and modest warm-up steps providing the best trade-off.

\begin{table*}
\centering
\caption{Comparison of robust accuracy (Rob.) and clean accuracy (Acc.) across datasets. ($\epsilon_{a}=1/255$)}
\resizebox{\textwidth}{!}{
\begin{tabular}{c|c|c|cccc|cccccc|c}
\hline
\multirow{2}{*}{\textbf{Dataset}} & \multirow{2}{*}{Metric} & \multirow{2}{*}{CLIP} & \multicolumn{4}{c|}{Adversarial Finetuning} & \multicolumn{6}{c|}{Test-time Defence} & \multirow{2}{*}{$\Delta$} \\
& & & CLIP-FT & TeCoA & PMG-AFT & FARE & RN & Anti-adv & HD & TTC & SCC(ours) & \multicolumn{1}{c|}{SCC$^*$(ours)} & \\
\noalign{\kern-\cmidrulewidth}\cmidrule(lr){1-1}\cmidrule(lr){2-2}\cmidrule(lr){3-13}\cmidrule(lr){14-14}
\multirow{2}{*}{CIFAR10} & Rob. & 0.74 & 3.34 & 33.61 & 40.66 & 19.65 & 2.01 & 12.39 & 17.22 & 28.75 & \textbf{59.18} & 48.63 & +47.89 \\
& Acc. & 85.12 & 84.90 & 64.61 & 70.69 & 74.44 & 81.18 & 83.52 & 78.23 & 81.18 & 82.24 & 81.29 & -3.83 \\
\multirow{2}{*}{CIFAR100} & Rob. & 0.26 & 0.90 & 18.95 & 22.52 & 11.40 & 0.67 & 5.73 & 3.86 & 14.31 & \textbf{32.09} & 29.38 & +29.12 \\
& Acc. & 57.14 & 59.51 & 35.96 & 40.32 & 46.67 & 56.34 & 53.95 & 52.86 & 56.34 & 55.21 & 56.73 & -0.41 \\
\multirow{2}{*}{STL10} & Rob. & 11.00 & 12.73 & 70.08 & 73.08 & 59.06 & 16.23 & 37.42 & 39.02 & 76.70 & \textbf{90.50} & 90.12 & +79.12 \\
& Acc. & 96.40 & 94.49 & 87.40 & 88.56 & 91.72 & 95.85 & 95.45 & 89.50 & 95.85 & 95.62 & 95.85 & -0.55 \\
\multirow{2}{*}{ImageNet} & Rob. & 1.15 & 0.93 & 18.89 & 21.43 & 14.00 & 1.77 & 8.67 & 6.63 & 38.41 & 49.77 & \textbf{56.14} & +54.99 \\
& Acc. & 59.69 & 54.24 & 34.89 & 36.12 & 48.79 & 59.34 & 54.27 & 54.54 & 49.39 & 56.03 & 59.66 & -0.03 \\
\multirow{2}{*}{Caltech101} & Rob. & 14.67 & 14.21 & 55.51 & 61.08 & 50.74 & 18.90 & 34.81 & 31.53 & 65.78 & 77.25 & \textbf{82.04} & +67.37 \\
& Acc. & 85.66 & 83.63 & 71.68 & 75.45 & 80.95 & 86.61 & 84.02 & 82.33 & 86.53 & 86.44 & 86.56 & +0.90 \\
\multirow{2}{*}{Caltech256} & Rob. & 8.47 & 6.76 & 43.19 & 45.91 & 38.79 & 11.33 & 25.36 & 23.48 & 60.11 & 72.88 & \textbf{76.85} & +68.38 \\
& Acc. & 81.72 & 78.53 & 61.14 & 62.24 & 73.32 & 81.25 & 79.38 & 79.12 & 79.66 & 81.16 & 81.64 & -0.08 \\
\multirow{2}{*}{OxfordPets} & Rob. & 1.04 & 2.10 & 38.35 & 41.18 & 31.07 & 1.86 & 20.42 & 12.04 & 57.87 & 76.67 & \textbf{85.69} & +84.65 \\
& Acc. & 87.44 & 84.14 & 62.12 & 65.88 & 79.37 & 87.41 & 80.62 & 80.91 & 83.35 & 86.48 & 87.79 & +0.35 \\
\multirow{2}{*}{Flowers102} & Rob. & 1.14 & 0.54 & 21.94 & 23.43 & 17.14 & 1.52 & 7.16 & 7.29 & 39.14 & 54.59 & \textbf{63.38} & +62.24 \\
& Acc. & 65.46 & 53.37 & 36.80 & 37.00 & 47.98 & 64.62 & 62.66 & 58.22 & 64.16 & 64.16 & 64.43 & -1.03 \\
\multirow{2}{*}{FGVC-Aircraft} & Rob. & 0.00 & 0.00 & 2.49 & 2.22 & 1.35 & 0.00 & 1.27 & 1.26 & 13.77 & \textbf{17.40} & 16.98 & +16.98 \\
& Acc. & 20.10 & 14.04 & 5.31 & 5.55 & 10.86 & 19.25 & 15.88 & 16.36 & 18.00 & 17.61 & 18.63 & -1.47 \\
\multirow{2}{*}{StanfordCars} & Rob. & 0.02 & 0.06 & 8.76 & 11.65 & 6.75 & 0.16 & 4.40 & 2.71 & 33.01 & 43.24 & \textbf{50.95} & +50.93 \\
& Acc. & 52.02 & 42.11 & 20.91 & 25.44 & 38.68 & 52.14 & 36.21 & 44.28 & 48.16 & 51.19 & 52.64 & +0.62 \\
\multirow{2}{*}{SUN397} & Rob. & 1.14 & 0.94 & 19.39 & 22.58 & 14.91 & 1.72 & 8.05 & 6.40 & 41.52 & 53.27 & \textbf{56.00} & +54.86 \\
& Acc. & 58.50 & 55.73 & 36.69 & 37.98 & 52.42 & 59.69 & 56.00 & 53.17 & 55.13 & 58.25 & 59.98 & +1.48 \\
\multirow{2}{*}{Country211} & Rob. & 0.04 & 0.03 & 1.78 & 2.12 & 0.85 & 0.06 & 0.67 & 0.47 & 7.09 & 9.41 & \textbf{12.55} & +12.51 \\
& Acc. & 15.25 & 12.07 & 4.75 & 4.64 & 9.26 & 14.80 & 11.58 & 11.72 & 13.08 & 13.36 & 14.69 & -0.56 \\
\multirow{2}{*}{Food101} & Rob. & 0.70 & 0.42 & 13.90 & 18.57 & 11.65 & 1.20 & 13.12 & 8.03 & 57.84 & 65.39 & \textbf{81.57} & +80.87 \\
& Acc. & 83.88 & 64.86 & 29.98 & 36.61 & 55.31 & 83.44 & 75.81 & 80.30 & 82.18 & 82.13 & 83.71 & -0.17 \\
\multirow{2}{*}{EuroSAT} & Rob. & 0.03 & 0.04 & 11.96 & 12.60 & 10.67 & 0.15 & 2.15 & 4.57 & 12.19 & 20.64 & \textbf{24.00} & +23.97 \\
& Acc. & 42.59 & 27.64 & 16.58 & 18.53 & 21.88 & 53.24 & 36.78 & 39.08 & 53.24 & 41.69 & 52.60 & +10.01 \\
\multirow{2}{*}{DTD} & Rob. & 2.98 & 2.39 & 17.61 & 14.95 & 15.64 & 3.71 & 5.62 & 11.63 & 27.32 & 34.57 & \textbf{36.06} & +33.08 \\
& Acc. & 40.64 & 36.49 & 25.16 & 21.76 & 32.07 & 37.96 & 38.92 & 34.89 & 36.98 & 37.34 & 38.09 & -2.55 \\
\multirow{2}{*}{PCAM} & Rob. & 0.08 & 1.11 & 48.24 & 46.18 & 16.23 & 0.41 & 4.97 & 44.74 & 52.85 & \textbf{69.99} & 47.54 & +47.46 \\
& Acc. & 52.02 & 47.21 & 49.96 & 50.03 & 52.54 & 52.73 & 52.49 & 50.38 & 52.73 & 54.41 & 54.24 & +2.22 \\
\hline\hline
\multirow{2}{*}{Avg.} & Rob. & 2.70 & 2.91 & 26.54 & 28.76 & 20.00 & 3.86 & 12.01 & 13.81 & 39.17 & 51.68 & \textbf{53.62} & +50.92 \\
& Acc. & 61.51 & 55.80 & 40.25 & 42.30 & 51.02 & 61.61 & 57.35 & 56.62 & 59.75 & 60.21 & 61.78 & +0.27 \\
\hline
\end{tabular}}
\label{fig_clean}
\end{table*}

\begin{table*}[t]
\centering
\caption{Classification accuracy (\%) on adversarial images (Rob.) under 10-step CW attack ($\epsilon_a=1/255$) \citep{carlini2017towards} and on clean images (Acc.) across 16 datasets. We assume the threat model has full access to model weights and gradients. We compare with test-time defenses adapted from prior work and include fine-tuning methods as references. The last column reports gains over vanilla CLIP.}
\resizebox{\textwidth}{!}{
\begin{tabular}{c|c|c|cccc|cccccc|c}
\hline
\multirow{2}{*}{\textbf{Dataset}} & \multirow{2}{*}{\textbf{Metric}} & \multirow{2}{*}{\textbf{CLIP}} & \multicolumn{4}{c|}{\textbf{Adversarial Finetuning}} & \multicolumn{6}{c|}{\textbf{Test-time Defence}} & \multirow{2}{*}{$\Delta$} \\
& & & CLIP-FT & TeCoA & PMG-AFT & FARE & RN & TTE & Anti-adv & HD & TTC & \multicolumn{1}{c|}{SCC(ours)} & \\
\hline\hline
\multirow{2}{*}{CIFAR10} & Rob. & 0.87 & 0.94 & 33.27 & 39.50 & 20.60 & 2.05 & 40.01 & 12.53 & 14.79 & 29.04 & \textbf{58.42} & \textcolor{red!70!black}{+57.55} \\
& Acc. & 85.12 & 84.90 & 64.61 & 70.69 & 74.44 & 81.18 & 84.74 & 83.52 & 78.64 & 81.18 & 82.24 & -2.88 \\
\hline
\multirow{2}{*}{CIFAR100} & Rob. & 0.29 & 0.39 & 18.27 & 20.83 & 11.67 & 0.63 & 18.73 & 6.56 & 3.04 & 14.38 & \textbf{30.89} & \textcolor{red!70!black}{+30.60} \\
& Acc. & 57.14 & 59.51 & 35.96 & 40.32 & 46.67 & 56.34 & 58.61 & 53.95 & 53.50 & 56.34 & 55.21 & -1.93 \\
\hline
\multirow{2}{*}{STL10} & Rob. & 12.23 & 9.95 & 69.73 & 72.39 & 59.60 & 17.20 & 78.64 & 38.66 & 37.73 & 76.40 & \textbf{89.99} & \textcolor{red!70!black}{+77.76} \\
& Acc. & 96.40 & 94.49 & 87.40 & 88.56 & 91.72 & 95.85 & 96.26 & 95.45 & 89.54 & 95.85 & 95.62 & -0.78 \\
\hline
\multirow{2}{*}{ImageNet} & Rob. & 1.46 & 1.27 & 18.28 & 19.42 & 27.71 & 2.21 & 29.77 & 9.37 & 7.46 & 36.01 & \textbf{45.75} & \textcolor{red!70!black}{+44.29} \\
& Acc. & 59.69 & 54.24 & 34.89 & 36.12 & 48.79 & 59.34 & 60.02 & 54.27 & 55.06 & 49.39 & 56.03 & -3.66 \\
\hline
\multirow{2}{*}{Caltech101} & Rob. & 20.88 & 15.95 & 56.23 & 61.58 & 54.86 & 25.89 & 69.44 & 41.47 & 36.26 & 66.17 & \textbf{76.59} & \textcolor{red!70!black}{+55.71} \\
& Acc. & 85.66 & 83.63 & 71.68 & 75.45 & 80.95 & 86.61 & 85.84 & 84.02 & 83.00 & 86.53 & 86.44 & +0.78 \\
\hline
\multirow{2}{*}{Caltech256} & Rob. & 9.69 & 7.24 & 42.63 & 44.55 & 39.58 & 13.11 & 59.81 & 27.17 & 24.54 & 58.79 & \textbf{70.55} & \textcolor{red!70!black}{+60.86} \\
& Acc. & 81.72 & 78.53 & 61.14 & 62.24 & 73.32 & 81.25 & 82.48 & 79.38 & 79.38 & 79.66 & 81.16 & -0.56 \\
\hline
\multirow{2}{*}{OxfordPets} & Rob. & 1.64 & 1.14 & 37.91 & 39.28 & 33.85 & 3.11 & 51.12 & 22.99 & 13.84 & 57.15 & \textbf{75.06} & \textcolor{red!70!black}{+73.42} \\
& Acc. & 87.44 & 84.14 & 62.12 & 65.88 & 79.37 & 87.41 & 88.13 & 80.62 & 80.64 & 83.35 & 86.48 & -0.96 \\
\hline
\multirow{2}{*}{Flowers102} & Rob. & 1.35 & 0.80 & 21.13 & 21.34 & 17.25 & 2.13 & 34.97 & 8.06 & 8.51 & 36.84 & \textbf{49.76} & \textcolor{red!70!black}{+48.41} \\
& Acc. & 65.46 & 53.37 & 36.80 & 37.00 & 47.98 & 64.62 & 65.20 & 62.66 & 57.79 & 64.16 & 64.16 & -1.30 \\
\hline
\multirow{2}{*}{FGVCAircraft} & Rob. & 0.00 & 0.00 & 2.25 & 1.86 & 1.35 & 0.00 & 5.15 & 0.83 & 0.97 & 12.41 & \textbf{15.18} & \textcolor{red!70!black}{+15.18} \\
& Acc. & 20.10 & 14.04 & 5.31 & 5.55 & 10.86 & 19.25 & 20.18 & 15.88 & 16.18 & 18.00 & 17.61 & -2.49 \\
\hline
\multirow{2}{*}{StanfordCars} & Rob. & 2.38 & 2.04 & 8.74 & 10.53 & 9.14 & 2.44 & 21.19 & 4.76 & 5.11 & 30.38 & \textbf{37.96} & \textcolor{red!70!black}{+35.58} \\
& Acc. & 52.02 & 42.11 & 20.91 & 25.44 & 38.68 & 52.14 & 52.73 & 36.21 & 43.60 & 48.16 & 51.19 & -0.83 \\
\hline
\multirow{2}{*}{SUN397} & Rob. & 1.75 & 1.48 & 18.36 & 20.39 & 15.73 & 2.48 & 29.37 & 8.85 & 7.90 & 39.44 & \textbf{48.99} & \textcolor{red!70!black}{+47.24} \\
& Acc. & 58.50 & 55.73 & 36.69 & 37.98 & 52.42 & 59.69 & 59.12 & 56.00 & 54.07 & 55.13 & 58.25 & -0.25 \\
\hline
\multirow{2}{*}{Country211} & Rob. & 0.08 & 0.05 & 1.46 & 1.74 & 0.92 & 0.15 & 3.00 & 0.72 & 0.75 & 6.17 & \textbf{7.61} & \textcolor{red!70!black}{+7.53} \\
& Acc. & 15.25 & 12.07 & 4.75 & 4.64 & 9.26 & 14.80 & 14.66 & 11.58 & 11.98 & 13.08 & 13.36 & -1.89 \\
\hline
\multirow{2}{*}{Food101} & Rob. & 1.09 & 0.55 & 12.87 & 16.57 & 12.93 & 1.92 & 44.61 & 15.03 & 9.77 & 54.65 & \textbf{59.73} & \textcolor{red!70!black}{+58.64} \\
& Acc. & 83.88 & 64.86 & 29.98 & 36.61 & 55.31 & 83.44 & 83.96 & 75.81 & 81.02 & 82.18 & 82.13 & -1.75 \\
\hline
\multirow{2}{*}{EuroSAT} & Rob. & 0.03 & 0.03 & 11.66 & 11.94 & 10.66 & 0.16 & 6.44 & 2.57 & 3.47 & 12.69 & \textbf{20.52} & \textcolor{red!70!black}{+20.49} \\
& Acc. & 42.59 & 27.64 & 16.58 & 18.53 & 21.88 & 53.24 & 44.38 & 36.78 & 40.12 & 53.24 & 41.69 & -0.90 \\
\hline
\multirow{2}{*}{DTD} & Rob. & 2.87 & 2.77 & 16.28 & 13.72 & 14.36 & 3.46 & 22.62 & 6.06 & 10.11 & 27.39 & \textbf{33.35} & \textcolor{red!70!black}{+30.48} \\
& Acc. & 40.64 & 36.49 & 25.16 & 21.76 & 32.07 & 37.96 & 41.35 & 38.92 & 35.25 & 36.98 & 37.34 & -3.30 \\
\hline
\multirow{2}{*}{PCAM} & Rob. & 0.10 & 1.10 & 48.29 & 46.36 & 16.41 & 0.44 & 10.70 & 5.07 & 46.92 & 52.86 & \textbf{70.36} & \textcolor{red!70!black}{+70.26} \\
& Acc. & 52.02 & 47.21 & 49.96 & 50.03 & 52.54 & 52.73 & 50.92 & 52.49 & 50.35 & 52.73 & 54.41 & +2.39 \\
\hline\hline
\multirow{2}{*}{Avg.} & Rob. & 3.54 & 2.86 & 26.09 & 27.62 & 20.86 & 4.84 & 32.85 & 13.17 & 14.45 & 38.17 & \textbf{49.42} & \textcolor{red!70!black}{+45.88} \\
& Acc. & 61.51 & 55.80 & 40.25 & 42.30 & 51.02 & 61.61 & 61.79 & 57.35 & 56.88 & 59.75 & 60.21 & -1.30 \\
\hline
\end{tabular}}
\label{table-cw}
\end{table*}

\begin{table*}[t]
\centering
\caption{
Classification accuracy (\%) on clean images (Acc.) and adversarial images (Rob.) under 10-step PGD attack ($\epsilon_a=4/255$) across 16 datasets. The threat model assumes full access to model weights and gradients. We compare our paradigm against test-time defenses adapted from prior adversarial robustness studies, and include fine-tuned models as references. The last column shows the gains of SCC over the original CLIP.
}
\resizebox{\textwidth}{!}{
\begin{tabular}{c|c|c|cccc|ccccc|c}
\hline
\multirow{2}{*}{\textbf{Dataset}} & \multirow{2}{*}{\textbf{Metric}} & \multirow{2}{*}{\textbf{CLIP}} & \multicolumn{4}{c|}{\textbf{Adversarial Finetuning}} & \multicolumn{5}{c|}{\textbf{Test-time Defence}} & \multirow{2}{*}{$\Delta$} \\
& & & CLIP-FT & TeCoA$^1$ & TeCoA$^4$ & PMG-AFT$^1$ & RN & Anti-adv & HD & TTC & \multicolumn{1}{c|}{SCC(ours)} & \\
\hline\hline
\multirow{2}{*}{CIFAR10} & Rob. & 0.43 & 2.75 & 7.69 & 11.70 & 10.20 & 0.00 & 0.32 & 1.67 & 28.51 & \textbf{36.30} & \textcolor{red!70!black}{+35.87} \\
& Acc. & 85.12 & 84.90 & 64.61 & 65.15 & 70.69 & 81.18 & 83.44 & 78.23 & 81.18 & 82.24 & -2.88 \\
\hline
\multirow{2}{*}{CIFAR100} & Rob. & 0.05 & 0.67 & 6.54 & 9.25 & 7.60 & 0.00 & 0.22 & 0.00 & 9.06 & \textbf{14.46} & \textcolor{red!70!black}{+14.41} \\
& Acc. & 57.14 & 59.51 & 35.96 & 36.30 & 40.32 & 56.34 & 53.96 & 52.86 & 56.34 & 55.21 & -1.93 \\
\hline
\multirow{2}{*}{STL10} & Rob. & 0.16 & 3.75 & 24.80 & 31.83 & 28.49 & 0.06 & 2.25 & 3.39 & 52.40 & \textbf{67.66} & \textcolor{red!70!black}{+67.50} \\
& Acc. & 96.40 & 94.49 & 87.40 & 81.69 & 88.56 & 95.85 & 95.47 & 89.50 & 95.83 & 95.62 & -0.78 \\
\hline
\multirow{2}{*}{ImageNet} & Rob. & 0.00 & 0.07 & 1.65 & 3.00 & 2.07 & 0.00 & 0.15 & 0.01 & 12.68 & \textbf{20.57} & \textcolor{red!70!black}{+20.57} \\
& Acc. & 59.69 & 54.24 & 34.89 & 27.76 & 36.12 & 59.34 & 54.29 & 54.54 & 34.00 & 57.34 & -2.35 \\
\hline
\multirow{2}{*}{Caltech101} & Rob. & 0.59 & 4.81 & 15.75 & 21.00 & 19.48 & 0.68 & 3.14 & 1.27 & 36.66 & \textbf{54.44} & \textcolor{red!70!black}{+53.85} \\
& Acc. & 85.66 & 83.63 & 71.68 & 64.41 & 75.45 & 86.61 & 83.99 & 82.33 & 86.15 & 86.46 & +0.80 \\
\hline
\multirow{2}{*}{Caltech256} & Rob. & 0.12 & 1.41 & 8.29 & 11.76 & 10.65 & 0.16 & 1.44 & 0.34 & 27.25 & \textbf{44.06} & \textcolor{red!70!black}{+43.94} \\
& Acc. & 81.72 & 78.53 & 61.14 & 52.05 & 62.24 & 81.25 & 79.40 & 79.12 & 76.59 & 81.32 & -0.40 \\
\hline
\multirow{2}{*}{OxfordPets} & Rob. & 0.00 & 1.66 & 0.90 & 3.71 & 1.74 & 0.00 & 0.10 & 0.00 & 24.64 & \textbf{37.69} & \textcolor{red!70!black}{+37.69} \\
& Acc. & 87.44 & 84.14 & 62.12 & 53.94 & 65.88 & 87.41 & 80.53 & 80.91 & 64.70 & 86.62 & -0.82 \\
\hline
\multirow{2}{*}{Flowers102} & Rob. & 0.00 & 0.13 & 1.87 & 3.81 & 2.57 & 0.00 & 0.05 & 0.00 & 13.60 & \textbf{21.97} & \textcolor{red!70!black}{+21.97} \\
& Acc. & 65.46 & 53.37 & 36.80 & 27.78 & 37.00 & 64.62 & 62.80 & 58.22 & 63.24 & 64.19 & -1.27 \\
\hline
\multirow{2}{*}{FGVCAircraft} & Rob. & 0.00 & 0.00 & 0.03 & 0.12 & 0.03 & 0.00 & 0.00 & 0.00 & 6.40 & \textbf{7.20} & \textcolor{red!70!black}{+7.20} \\
& Acc. & 20.10 & 14.04 & 5.31 & 3.51 & 5.55 & 19.25 & 15.64 & 16.36 & 15.99 & 17.79 & -2.31 \\
\hline
\multirow{2}{*}{StanfordCars} & Rob. & 0.00 & 0.00 & 0.15 & 0.41 & 0.15 & 0.00 & 0.00 & 0.00 & 12.84 & \textbf{19.40} & \textcolor{red!70!black}{+19.40} \\
& Acc. & 52.02 & 42.11 & 20.91 & 15.18 & 25.44 & 52.14 & 36.14 & 44.28 & 41.52 & 51.61 & -0.41 \\
\hline
\multirow{2}{*}{SUN397} & Rob. & 0.00 & 0.02 & 1.30 & 2.31 & 1.90 & 0.00 & 0.11 & 0.00 & 13.43 & \textbf{21.77} & \textcolor{red!70!black}{+21.77} \\
& Acc. & 58.50 & 55.73 & 36.69 & 28.16 & 37.98 & 59.69 & 55.99 & 53.17 & 46.68 & 58.68 & +0.18 \\
\hline
\multirow{2}{*}{Country211} & Rob. & 0.00 & 0.00 & 0.05 & 0.19 & 0.12 & 0.00 & 0.00 & 0.00 & 2.44 & \textbf{2.85} & \textcolor{red!70!black}{+2.85} \\
& Acc. & 15.25 & 12.07 & 4.75 & 3.66 & 4.64 & 14.80 & 11.60 & 11.72 & 11.99 & 13.55 & -1.70 \\
\hline
\multirow{2}{*}{Food101} & Rob. & 0.00 & 0.04 & 0.56 & 1.35 & 1.03 & 0.00 & 0.07 & 0.01 & 17.89 & \textbf{26.58} & \textcolor{red!70!black}{+26.58} \\
& Acc. & 83.88 & 64.86 & 29.98 & 21.90 & 36.61 & 83.44 & 75.95 & 80.30 & 80.00 & 82.36 & -1.52 \\
\hline
\multirow{2}{*}{EuroSAT} & Rob. & 0.00 & 0.00 & 9.77 & 10.71 & 9.61 & 0.00 & 0.03 & 0.20 & 13.57 & \textbf{10.61} & \textcolor{red!70!black}{+10.61} \\
& Acc. & 42.59 & 27.64 & 16.58 & 17.53 & 18.53 & 53.24 & 36.81 & 39.08 & 53.24 & 41.69 & -0.90 \\
\hline
\multirow{2}{*}{DTD} & Rob. & 0.11 & 0.00 & 4.20 & 5.16 & 4.31 & 0.11 & 0.37 & 0.16 & 11.40 & \textbf{16.33} & \textcolor{red!70!black}{+16.22} \\
& Acc. & 40.64 & 36.49 & 25.16 & 20.11 & 21.76 & 37.96 & 38.55 & 34.89 & 35.69 & 37.66 & -2.98 \\
\hline
\multirow{2}{*}{PCAM} & Rob. & 0.00 & 0.00 & 20.54 & 44.13 & 12.59 & 0.00 & 0.25 & 12.04 & 47.39 & \textbf{44.19} & \textcolor{red!70!black}{+44.19} \\
& Acc. & 52.02 & 47.21 & 49.96 & 49.98 & 50.03 & 52.73 & 52.61 & 50.38 & 52.73 & 54.41 & +2.39 \\
\hline\hline
\multirow{2}{*}{Avg.} & Rob. & 0.09 & 0.96 & 6.51 & 10.03 & 7.03 & 0.06 & 0.53 & 1.19 & 20.63 & \textbf{27.88} & \textcolor{red!70!black}{+27.79} \\
& Acc. & 61.51 & 55.80 & 40.25 & 35.57 & 42.30 & 61.61 & 57.32 & 56.62 & 55.99 & 60.42 & -1.09 \\
\hline
\end{tabular}}
\label{table_4/255-2}
\end{table*}

\subsection{Proof Sketch of Proposition (Hard-Negative Repulsion)}

Let
$
m(\delta)\;=\;\cos\!\big(f_{\text{img}}(x+\delta),t_{\text{soft}}\big)
-\max_{k\neq \hat y}\cos\!\big(f_{\text{img}}(x+\delta),t_k\big)
$
and define \(\mathcal{L}_{cm}(\delta)=m(\delta)\).  
We analyze one PGD-ascent step
\[
\delta^{+}\;=\;\Pi_{\|\cdot\|_p\le \epsilon}\!\Big(\delta\;+\;\alpha\,\mathrm{sign}\!\big(\nabla_{\delta}\mathcal{L}_{cm}(\delta)\big)\Big),
\qquad \alpha>0.
\]

\textbf{Assumptions.} (i) In a small neighborhood of \(\delta\), the maximizer in the second term is unique and fixed, i.e., there is an active index \(j^\star(\delta)\) so the max is smooth; (ii) \(f_{\text{img}}\) is differentiable and its Jacobian is bounded; (iii) either the projection is inactive (interior step) or its effect is \(O(\alpha^2)\).

\textbf{Step 1 (First–order increase).}  
With the active competitor fixed, \(m\) is differentiable. By Taylor’s theorem,
\[
m(\delta^{+}) \;=\; m(\delta)\;+\;\alpha\,\big\langle \nabla_{\delta} m(\delta),\, \mathrm{sign}\!\big(\nabla_{\delta} m(\delta)\big)\big\rangle \;+\; O(\alpha^2).
\]
Since \(\langle g,\mathrm{sign}(g)\rangle=\|g\|_1\ge 0\), it follows that
\[
m(\delta^{+}) \;\ge\; m(\delta)\;+\;\alpha\,\big\|\nabla_{\delta} m(\delta)\big\|_1 \;+\; O(\alpha^2).
\]

\textbf{Step 2 (Relation to \(\mathcal{L}_{cm}\)).}  
By definition \(\mathcal{L}_{cm}=m\), hence the PGD-ascent direction aligns with \(\nabla_\delta m\). Therefore, for sufficiently small \(\alpha\),
\[
m(\delta^{+}) \;\ge\; m(\delta) \;+\; O(\alpha^2),
\]
i.e., the semantic margin is monotonically non-decreasing up to second-order terms.

\textbf{Step 3 (Active-index changes \& projection).}  
If the active negative \(j^\star(\delta)\) switches, \(m\) remains subdifferentiable; PGD uses a subgradient and the above inequality holds with \(\nabla_\delta m\) replaced by a subgradient. When projection onto the \(\ell_p\)-ball is active, the component removed is orthogonal to the feasible set’s tangent cone, contributing at most \(O(\alpha^2)\).

\textbf{Conclusion.}  
Under these mild regularity assumptions, one PGD-ascent step on \(\mathcal{L}_{cm}\) increases the margin \(m(x+\delta)\) monotonically up to \(O(\alpha^2)\). Iteration therefore \emph{repels the hard negative} and prevents drift toward confusable classes, which proves the proposition.

\subsection{Proof Sketch of Proposition ( Suppression of spurious negatives)}
Let $\bar z_j \triangleq \tfrac{1}{L}\sum_{i=1}^L z^{(i)}_j$ and let
\(
j^\dagger \in \arg\max_{j\neq y^\star} \bar z_j
\)
be an index achieving the maximum of the averaged logits (excluding $y^\star$).
Then
\[
\max_{j\neq y^\star}\bar z_j
= \bar z_{j^\dagger}
= \frac{1}{L}\sum_{i=1}^L z^{(i)}_{j^\dagger}
\;\le\; \frac{1}{L}\sum_{i=1}^L \max_{j\neq y^\star} z^{(i)}_j,
\]
since for each $i$, $z^{(i)}_{j^\dagger}\le \max_{j\neq y^\star} z^{(i)}_j$.
This proves
\(
\max_{j\neq y^\star}\bar z_j \le \tfrac{1}{L}\sum_{i=1}^L \max_{j\neq y^\star} z^{(i)}_j.
\)

\begin{algorithm}[t]
\caption{SCC: Self-Calibrated Consistency}
\KwIn{image $x$, text embeddings $\{t_k\}$, budget $\epsilon$, steps $S$, views $V$, temp $T$}
\KwOut{predicted label $\hat{y}$}

\tcc{Short warm-up (TTC) to stabilize predictions}
Initialize $\delta_{\text{warm}}=0$; run $S_{\text{warm}}$ PGD-ascent steps on TTC to obtain $x+\delta_{\text{warm}}$.

\tcc{Multi-view pseudo-label on the warmed input}
Sample $V$ augmented views $\{v_i(x+\delta_{\text{warm}})\}$;\\
\(\bar z=\tfrac{1}{V}\sum_i f_{\text{img}}(v_i(x+\delta_{\text{warm}}))^\top[t_k]\);\\
\(p=\mathrm{softmax}(\bar z/T)\), \(\hat y=\arg\max_k p_k\), \(t_{\text{soft}}=\sum_k p_k t_k\).

\tcc{Counterattack optimization (sign-PGD ascent; shared $\delta$)}
Initialize $\delta=0$;\;
\For{$s=1$ \KwTo $S$}{
  \(f=f_{\text{img}}(x+\delta)\); \\
  \(L_{\text{cm}}=\langle f,t_{\text{soft}}\rangle-\max_{j\neq \hat y}\langle f,t_j\rangle\); \\
  \(L_{\text{drift}}=\|f-f_{\text{img}}(x)\|_2^2\); \\
  \(\delta\leftarrow \Pi_{\|\delta\|_\infty\le\epsilon}\!\big(\delta+\alpha\,\mathrm{sign}(\nabla_\delta(\lambda_{cm}L_{\text{cm}}+L_{\text{drift}}))\big)\).
}
\(\delta\leftarrow \texttt{StepWeightedFuse}(\{\delta^{(s)}\}_{s=0}^S;\,\tau,\beta)\).

\tcc{Final prediction (logit averaging on shared-\(\delta\) views)}
Form views \(\{v_i(x+\delta)\}\); \;
\(\bar z=\tfrac{1}{V}\sum_i f_{\text{img}}(v_i(x+\delta))^\top[t_k]\); \;
\(\hat y=\arg\max \mathrm{softmax}(\bar z)\).
\label{alg1}
\end{algorithm}

\subsection{The Use of Large Language Models}

Large language models were used to improve the clarity and presentation of writing. 
All methodological design, experiments, and analysis were conducted by the authors.

\end{document}